\documentclass[3p,times]{elsarticle}

\usepackage{amssymb}
\usepackage{amsmath}
\usepackage{pdflscape}
\usepackage{lineno}
\usepackage{hyperref}
\usepackage{url}
\usepackage{multirow}
\usepackage{footnote}
\usepackage{doi}
\biboptions{numbers,sort&compress}
\usepackage[multiple]{footmisc}
\usepackage[ruled,vlined]{algorithm2e}
\DeclareMathOperator*{\argmax}{argmax}
\SetKwComment{Comment}{$\triangleright$\ }{}
\journal{\textbf{arXiv}}

\begin{document}
\biboptions{sort&compress}

\begin{frontmatter}

\title{TopicBERT: A Transformer transfer learning based memory--graph approach for multimodal streaming social media topic detection}

\author[prof]{Meysam Asgari-Chenaghlu}
\author[prof]{Mohammad-Reza Feizi-Derakhshi}
\author[prof]{Leili farzinvash}
\author[prof]{Mohammad-Ali Balafar}
\author[prof1]{Cina Motamed}

\address[prof]{Department of Computer Engineering, Faculty of Electrical and Computer Engineering, University of Tabriz, Iran}
\address[prof1]{Laboratoire D’Informatique Signal Et Image de la Côte D’Opale, Université du Littoral Côte D’Opale, Calais Cedex, France}

\begin{abstract}
Real time nature of social networks with bursty short messages and their respective large data scale spread among vast variety of topics are research interest of many researchers. These properties of social networks which are known as 5'Vs of big data has led to many unique and enlightenment algorithms and techniques applied to large social networking datasets and data streams. Many of these researches are based on detection and tracking of hot topics and trending social media events that help revealing many unanswered questions. These algorithms and in some cases software products mostly rely on the nature of the language itself. Although, other techniques such as unsupervised data mining methods are language independent but many requirements for a comprehensive solution are not met. Many research issues such as noisy sentences that adverse grammar and new online user invented words are challenging maintenance of a good social network topic detection and tracking methodology; The semantic relationship between words and in most cases, synonyms are also ignored by many of these researches. In this research, we use Transformers combined with an incremental community detection algorithm. Transformer in one hand, provides the semantic relation between words in different contexts. On the other hand, the proposed graph mining technique enhances the resulting topics with aid of simple structural rules. Named entity recognition from multimodal data, image and text, labels the named entities with entity type and the extracted topics are tuned using them. All operations of proposed system has been applied with big social data perspective under NoSQL technologies. In order to present a working and systematic solution, we combined MongoDB with Neo4j as two major database systems of our work. The proposed system shows higher precision and recall compared to other methods in three different datasets.
\end{abstract}


\end{frontmatter}

\newtheorem{definition}{Definition}
\section{Introduction}
\label{sec:introduction}
Internet and its applications are widely advancing towards social networks in which the user generated content plays a crucial role \cite{Dallas2012}. This content which is mostly generated by users, companies, news reporting agencies and etc., consists of text, image and video and has noise naturally \cite{Atefeh2015,Kaplan2010}. This media forms the foundation of social media delivery systems such as Facebook, Youtube, Instagram, Twitter, and etc. with different views and media orientations. Some of these social networks tend to be an online social TV such as Youtube. Content of these networks are generated in video format by users and posted in channels and communication between users is possible by commenting under these videos. Instagram was started as an image and short video sharing social network and the same communication method has been implemented. Facebook as another big social network, does not restrict users to post specific media types and also allows users to directly communicate with each other using direct chat which is recently available on Instagram too.

Among these popular social media delivery systems, Twitter has gained much attention in recent years due to its simplicity and user friendly interface. It is also known as the best social media for discovering ``\textit{what is happening right now}''\cite{Twitter2018}. Twitter is easily accessible by almost any device connected to Internet through a web browser or a third-party/official application. The nature of twitter short posts that are known as tweets has restrictions in size and at first it was supposed to be a gentle replacement for SMS \cite{Carlson2011}. In other words, twitter is the best social media for discovering news and real world events \cite{Kwak2010}. 

Reports show that on a daily basis, 500 million new tweets and on a monthly basis, 300 new user sign-ups happened on twitter in 2013; In 2018, it is estimated that 700 million tweets will be posted each day and growth rate for number of tweets per year to be around 32\% on 2013 \cite{TwitterStats}. These statistics show that users tend to produce large volumes of tweets on variety of types with high velocity and diverse veracity with different values; Mentioned characteristics of twitter streaming media takes it into account of social big data with respect to 5'V model \cite{Bello-Orgaz2016}.

Twitter has very rich data in form of tweets which can be accessed using Twitter API. This API made researchers capable of investigating its social big data in a streaming way or in a query based data retrieval fashion \cite{API2018}. Documentation about the API itself is available at twitter developer website\footnote{\url{https://developer.twitter.com/en}}. This API allows direct messaging, search, advertisement management, and account activity control. It has some restriction to prevent developers from abusing the service. For example, it has rate limit for user or application\footnote{\url{https://developer.twitter.com/en/docs/basics/rate-limiting}}.

This open and free dataset has been used by many researchers. Many advancements has been conducted in multiple research areas related to this social big data such as opinion mining, topic detection and tracking, user modeling, sentiment analysis, and etc. But the underlying conceptual perspective of researcher to the problem makes different outcomes with diverse real world applications. If the data is supposed to be treated as streaming social big data (as it is) then the problem would likely to be solved in an unsupervised or semi-supervised methodology. Some other researches try to solve the problem with respect to a static corpus but others tend to do so in an streaming big social data respect \cite{atefeh2015survey}.

On the other hand, many of the previous works assumed the correctness of incoming data for granted which is not true in many cases. For example, in the case of analyzing tweets related to regular users, many noisy content such as mistypes and grammar errors exist while in the case of official tweets like a news agency tweet, this problem can be ignored. The most missing part of this analysis is related to images and videos of the tweets. To the best of our knowledge, most of the existing methods did not use multimodal data to detect and track topics.

To overcome  the mentioned problems, we present a novel approach for topic detection and tracking problem. Our work consists of three major parts: multimodal named entity recognizer, memory graph, and Transformer based semantic similarity assignment. The combination of these three parts provides the final results. Orientation of presented study is as follows: Section \ref{sec:relatedWork} covers related works of the problem at hand. Section \ref{sec:topicbert} presents the proposed methodology for detecting and tracking topics. The experimental results at Section \ref{sec:experimentalResults} investigates the performance of our work on three distinct datasets. Finally, Section \ref{sec:conclusions} concludes the whole paper.

\section{Related work}
\label{sec:relatedWork}
This section describes various techniques proposed for topic detection and tracking from twitter. For reader curiosity and further information, Table \ref{tab:tableRelatedWorks} is inserted to cover most of related works in literature of TDT task on twitter.




The first methods we investigate here, are using frequent pattern mining as building block. The frequent pattern mining problem is a well studied area by virtue of its various applications in multiple domains like clustering and classification. This problem is initiated as finding relationships between items in a transaction database \cite{aggarwal2014frequent}, where the formal definition is given below:

\begin{definition}
	\label{def:fpm}
	Given a database $D$ consisting transactions $T_1, \dots, T_2$, determination of all possible patterns $P$ with at least $s$ frequency over all of transactions, is known as Frequent Pattern Mining.
\end{definition}

With respect to this definition, a document collection can be noted as a database of transactions. Depending on this technical terminology, a frequent pattern mining approach applied to collection of documents discovers relationships between words with frequency of at least $s$ \cite{ibrahim2017tools,atefeh2015survey}. To have a better formalization and homogeneity, just like a database of transactions, we use notation $D$ for collection of documents and $P$ for possible patterns which are relationships between words. 

Abstract topic on the other hand, is defined in the following:

\begin{definition}
	\label{def:topic}
	Set of abstract topics (or shortly $topics$) $\Theta$ in collection of documents $D$ are bag of words $W_\Theta$ related to each other by occurrence in the same documents with high frequency.
\end{definition}

Summing up definitions \ref{def:fpm} and \ref{def:topic}, topic detection task from a series of documents is almost equivalent to frequent pattern mining in database of transactions if the semantic relation between words is discarded from problem.

A major topic such as $\theta_i \in \Theta$ usually consists of multiple minor topics like $\theta_j$, based on this assumption, a minor topic is defined as: $\theta_j = \{W_{\theta_j} \mid W_{\theta_j} \subset W_{\theta_i} \}$. With association rule mining in mind \cite{zhang2002association}, statement $\theta_j \subset \theta_i$ holds true while the confidence $c$ for a minor topic depends on detection algorithm but at any means is less than one.

On the contrary, a real life event, or event for short is defined as \cite{atefeh2015survey}:

\begin{definition}
	\label{def:event}
	An event $e_i \in E$ is a real-life topic that occurs in real life at specific time period $\tau$.
\end{definition}

According to this definition, any event is also a topic but a topic $\theta_i$ needs to happen in real life to be counted as an event $e$. For example, the \texttt{Bongo Cat}\footnote{Bongo cat is a set of tweets identified by \texttt{\#BongoCat} hashtag: \url{https://www.twitter.com/hashtag/Bongocat}} is not a real life event but it is surely a topic. For event detection tasks, it is crucial to have more information about topics rather than depending only on documents; online sources such as Wikipedia\footnote{Wikipedia is a free encyclopedia, written by the people who use it: \url{https://www.wikipedia.org/}} are very helpful for information gathering on this specific problem.

These two problems, namely frequent pattern finding and topic detection, are dealt by a similar way. Frequent pattern mining has been a common way to solve the topic detection task over twitter data stream and some variants of formal methods has been proposed by researchers in \cite{Petkos2014,HuangJiajiaandPeng2015,Gaglio2015,Choi2019}.

In \cite{Petkos2014}, authors designed a soft frequent pattern mining approach to overcome the topic detection problem. This research aims at finding word co-occurrence large than two, in order to do so, likelihood of each word is computed separately. Equation \ref{eq:likelihood} shows likelihood of appearance for each word.

\begin{equation}
\label{eq:likelihood}
p(w|corpus) = \frac{N_w + \sigma}{(\sum_{u}^{n}N_u) + \sigma n}
\end{equation}

\noindent where $N_w$ is the number of occurrences of word $w$, and $\sigma$ equals to $0.5$. Denominator part in this equation is used to normalize the fraction. Grow rate of each word in corpus is computed to indicate the most discussed words.

After finding top-K frequent words, a co-occurrence vector is formed to add co-occurring words to the top-K word vectors. This expansion forms the foundation of soft frequent pattern mining algorithm (SFPM) utilized in \cite{Petkos2014} to solve the topic detection problem. Results presented in main article shows superiority of this method to $LDA$ \cite{Sayyadi2009} and a $Graph-based$ approach \cite{Quercia2012}, which is assumed to be the baseline for this research.

Like previous method, in \cite{Gaglio2015} a named entity recognition weight has been added to boost scores for each word. This simple addition to original algorithm (SFPM) improved topic recall, keyword precision and keyword recall on two different datasets.

Recently, high utility pattern mining (HUPM) has been introduced \cite{liu2012mining}. This approach for finding frequent item-sets with higher utility in twitter topic detection domain has been investigated in \cite{HuangJiajiaandPeng2015,Choi2019}. The proposed high utility pattern clustering in \cite{HuangJiajiaandPeng2015} considered the utility of each pattern in order to extract high utility patterns, afterwards it takes these patterns and uses an incremental pattern clustering scheme. $KNN$ and $modularity-based$ clustering employed by authors, try to identify new emerging patterns and coherent topics simultaneously. Finally, a group of words for each cluster of patterns are extracted to represent the topic of that specific cluster. All of mentioned steps together form HUPC framework (High Utility Pattern Clustering): text stream, text segments, top-K HUP mining, term indexing, HUP clustering, and post-processing.

Like two previous methods, proposed scheme in \cite{Choi2019} considered the problem of topic detection as a pattern mining problem and introduces a HUP mining algorithm in order to extract final topics. Internal, external, and transaction weighted utility of words has been used by authors to determine the utility of words. Minimum utility of each chunk of tweets is determined automatically by their algorithm. Also, for post-processing step, a $Topic-tree$ was proposed to extract actual topics from candidate ones. Experimental results shows superiority of this work compared to five different methods.

Some research \cite{saeed2018text,saeed2019event,saeed2019enhanced,saeed2020evesense} used dynamic heart beat graph technique (DHG) for streaming data on Twitter. The proposed algorithms construct a sub-graph for each sentence. These bi-clique sub-graphs are added to the main graph and a graph operation on the whole graph is performed based on the frequency of co-occuring words (edges between nodes). Their proposed method shows a novel approach on how graph methods can be applied to streaming text but on the other hand their work lacks multimodality and comprehensiveness.

However, in most of previous works the main contribution of authors is extracting topics with higher precision and recall, but none of these works except \cite{Choi2019,saeed2019enhanced} tried to take performance metrics in mind. An online topic detection and tracking scheme must fit in Definition \ref{def:NED} to be considered as a practical solution.



\begin{definition}
	\label{def:NED}
	New event detection (NED) is discovering new events from live stream $L$ in real time with no delay or at least with an acceptable delay $t_d$.
\end{definition}

If the extraction and tracking method is applied to collection of offline documents, it will be considered a RED methodology, which completely differs from NED. Definition \ref{def:RED} is a formalization of RED.

\begin{definition}
	\label{def:RED}
	Retrospective event detection (RED) is discovering previously unidentified events from collection of documents.
\end{definition}

Definitions \ref{def:NED} and \ref{def:RED} make a fair clarification of Table \ref{tab:tableRelatedWorks} and highlight differences among diverse approaches. In Table \ref{tab:tableRelatedWorks} we provide an overview of related works in this field and categorized them based on the NED and RED terminology.


\section{TopicBERT}
\label{sec:topicbert}
The three pillars of our proposed system (TopicBERT), namely, Transformer semantic similarity, word community detection, and multimodal named entity recognizer, are bounded together as it is presented in Figure \ref{fig:proposed_system}. The incoming data from social media stream such as Twitter is streamed into the MongoDB that feeds multimodal data into different elements of the system. In this design, MongoDB acts as a cache for whole system in the case of any long delay or failure in any system parts. Delays happen because different parts have different complexity and speed metrics. On the other hand, the neo4j\footnote{\url{https://neo4j.com/}} database is the repository of word graph in form of a memory graph. The following sections describe each part in more detail and the last section reviews how the whole system works in a harmony.
\subsection{Memory-Graph}
\label{sec:wordSubgraphMining}
\begin{figure}
	\centering
	\includegraphics[width=0.6\linewidth]{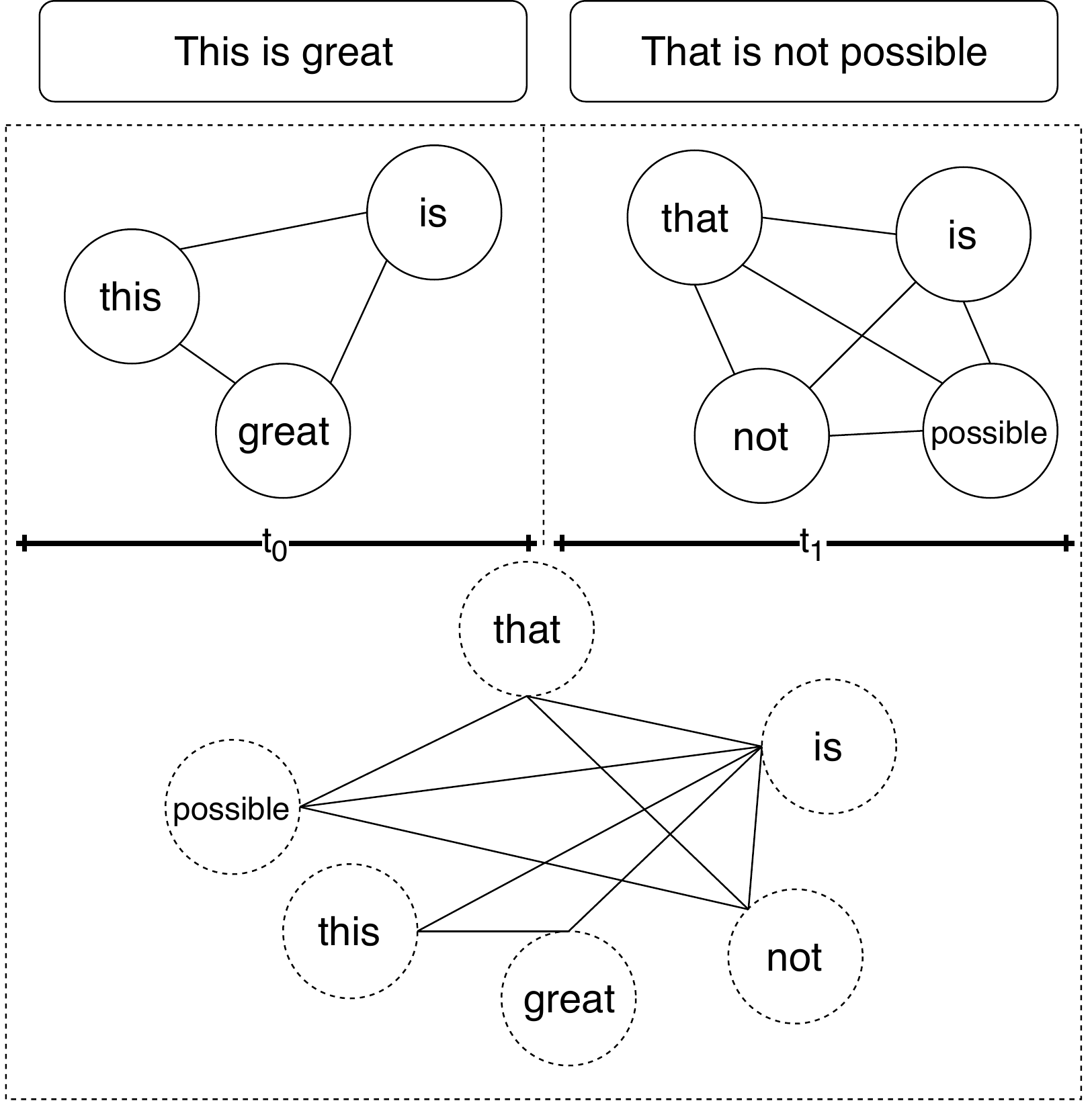}
	\caption{Word Graph formation example}
	\label{fig:wordgraph}
\end{figure}

Our approach identifies any document as a word graph such as $G_i=(V_i,E_i)$, in which the nodes represent words, and vertices are the co-occurrence relation of these nodes. A group of word graphs arriving in a streaming fashion form a larger graph such as $G=(V,E)$. Figure \ref{fig:wordgraph} shows an example of such graphs. Preprocessing tasks like tokenization and stemming can be applied before forming word graphs. Time slots are used to identify arrival of each document and accordingly its addition to $G$. Upon arrival of each separate and independent $G_i$, the cumulative word graph $G$ will be updated and each new node and edge will be added. Putting previous assumption in mind, each word graph has a time slot identifier label like $t$ and must be noted as $G_{(i,t)}$. Discarding duplicated words in each document yields to no further change in $G_{(i,t)}$ but the $G$ must be a weighted graph to uphold weight of relations. In our case, we use similarity metric for computing comulative weights that is explained in Section \ref{sec:transformer}.

Evolving nature of cumulative word graph $G$ indicates that streaming documents tend to complete it. In order to have a fading effect on $G$, we also consider it as a memory-graph (MG) that can remember and forget. A control parameter $\varrho$, controls the fading effect on memory graph, meaning, each node and edge would fade in MG as time passes (new word graphs arriving).

Forgetting and remembering properties of MG makes it reliable for extracting frequent patterns as relations that have certain properties in terms of MG's memory representation. In other words, relations that withstand the evolution of MG are those that are being repeatedly added. Weights of MG show the importance of particular relations over time, if any weight goes under a specific value, then the relation does not seem to be an important one anymore. If the weight shows a spiking growth, it means that the relation has been repeated with a high frequency in a short time. This property indicates an emerging relation if it disappears soon, otherwise, it will be assumed to be a \textit{common relation}.

All that is said draws a general assumption of MG which is core of current research to control the size of graph. Setting up forgetting rate can be done in diverse setups ranging from computational or cognitive approaches. In order to adapt a cognitive configuration bound with our presented MG, we implement \textbf{forgetting curve} \cite{wozniak1995two}. According to \textit{Hermann Ebbinghaus}'s work, who pioneered experimental study of memory, the forgetting curve is formulated as:

\begin{equation}\label{eq:forgettingcurve}
R = e^{-\frac{t}{\varrho}}
\end{equation}

\noindent where $R$ is retrievability, $\varrho$ is stability of memory and $t$ is time. $\varrho$ determines how fast retrievability falls over time in absence of any \textbf{memory recall}\footnote{\textbf{memory recall} is different from \textbf{recall}}.

Memory recall is defined as simply trying to remember items. However, there are two types of memory recall, \textbf{free recall} and \textbf{serial recall}. Free recall is an attempt to recall individual items regardless of their order while serial recall is attempting to recall the list of items in the ordered study. \textbf{Overlearning effect} is practicing memorization beyond required repetitions. This effect ensures that the memorized information will be more resistant to disruption or loss.

Using this effect and multiplying it to any values gives the fading effect over it. For the $t$ value in this equation, we will use the time that specific value last updated. If it has been a long time from its last update then the effect is high and if it goes under a specific threshold it will be completely forgotten (deleted).

However, the remembering and forgetting has its shortcomings too. For example, some words come from different contexts which need to be updated with different weights rather than just using +1 for any edge. On the other hand, words or phrases must have different importance because of their semantic nature; For example, a phrase related to a named entity such as person or location must be more important compared to a typical word. To enhance the described memory graph, subsection \ref{sec:transformer} and \ref{sec:multimodalner} describe our solutions.

\subsection{Transformer masked language modelling}
\label{sec:transformer}
Capturing the words and recording them in a graph form is explored in many previous works too. However, none of them tried to consider the semantic relation between words that appear in previous different contexts. For example, consider a word such as $W$, previously appearing in three different tweets of $T_1,T_2,T_3$ and the word $W'$ in only one tweet $T_4$, each of these two words had been on different contexts before, and no relation between them is available in our word graph. A new tweet arrives that has both of these words in the same context, a tweet such as $T_5$. The resulting sub-graph contains both of $W$ and $W'$ now and according to previous methods there should be an edge with the weight of one between the two words. Apart from the consideration that this assumption is wrong, in this section we try to employ Transformers to improve it.

We use the BERT transformer pretrained model\footnote{Available at: \url{https://storage.googleapis.com/bert_models/2020_02_20/uncased_L-4_H-512_A-8.zip}}\footnote{Source code from google: \url{https://github.com/google-research/bert}}\footnote{SentBert: \url{https://pypi.org/project/sentence-transformers/}} to compute the embedding for each incoming tweet and record it as an attribute for each node (word). The original version of BERT is not sufficient because it does not capture sentence similarity, instead we used a pre-trained version called SentBERT \cite{reimers2019sentence}. If another tweet comes that has the word, embedding attribute of node is updated using a weighted approach between previous and new embedding. In the case of computing score for two different nodes in one tweet, we use frequency and similarity score. Frequency is the number of occurrence of two nodes in the same tweets, and score is calculated using \eqref{eq:score}.

\begin{equation}\label{eq:score}
S_{i,j}^{t} = S_{i,j}^{t-1}+ cos(M_i^{t},M_j^{t})\text{, if } W_i \text{ and }W_j \in tweet_t
\end{equation}
\begin{equation}\label{eq:embedding}
M_i^{t} = \alpha M_i^{t-1}+ (1-\alpha) M_i^{tweet_t}\text{, if }W_i \in tweet_t
\end{equation}
where, $tweet_t$ is the tweet that arrives at time $t$, $S_{i.j}^t$ is the score between words $W_i$ and $W_j$, $M_i^t$  is the embedding of word $W_i$, $\alpha$ is discount value for aggregating previous embedding and the new one (in the range of $(0,1)$), and $t$ is the time.
According to \eqref{eq:embedding}, at each time a new tweet arrives and the embedding of nodes are updated with embedding provided from BERT (i.e., \texttt{[CLS]} token). The reason we have used \texttt{[CLS]} instead of each node embdding is that the context itself is more important than the word embedding and some words are tokenized into multiple tokens rather than one token which makes it harder to aggregate.
Scoring equation shown in \eqref{eq:score} provides a mechanism to score edges rather than naively using frequencies. This score equation at its best is equal to co-occurrence frequency, because larger values for $cos(M_i^{t},M_j^{t})$ at each step equals means higher context similarity.
However, this mechanism provides a smoother approach compared to frequency counting, but the online learning for fine tuning the underlying BERT language model has an undeniable importance. This fine tuning makes the language model more tuned to the task which the language is in form of tweets rather than formal essays or grammatically perfect articles. We adopt this assumption too, and use it in our setup to finetune the BERT at each step we gather batches of tweets. Utilizing MongoDB as a cache is important in this case. Additionally, if there is more than one server available, a dedicated server can be used for language modeling.

\subsection{Multimodal Named Entity Recognition}
\label{sec:multimodalner}
\begin{figure*}[!ht]
	\centering
	\includegraphics[width=\textwidth]{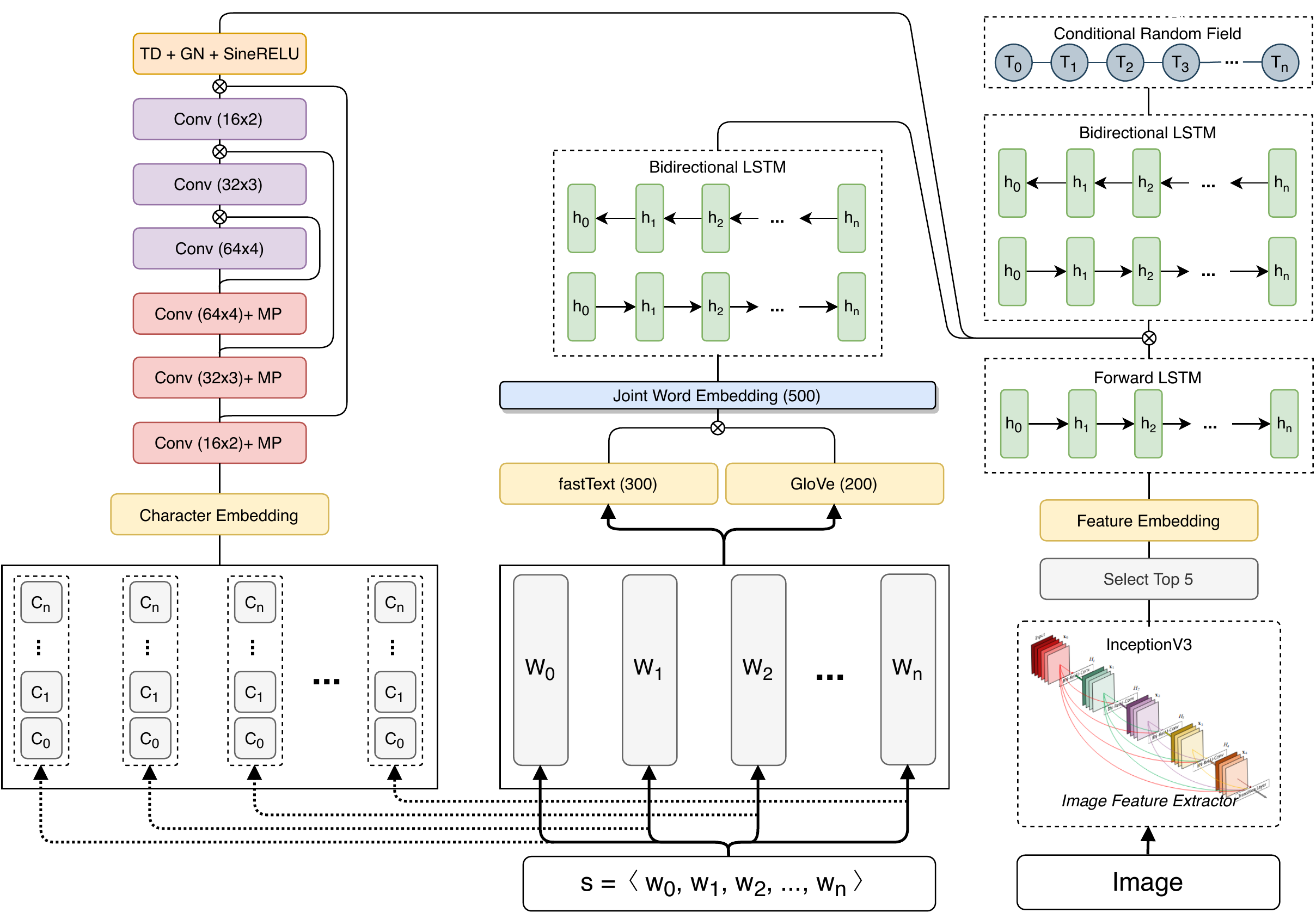}
	\caption{Multimodal NER model, courtesy of \cite{asgari2020multimodal}.}
	\label{fig:multimodalner}
\end{figure*}

Categorization of items in the graph, based on their types provides key features to highlight the node significance. For example, in our study, words or phrases that point to some named entity such as a person is more important in forming a topic. To find the named entities, we use the deep multimodal model described in \cite{asgari2020multimodal}.
Figure \ref{fig:multimodalner} shows the multimodal named entity recognizer model that utilizes both image and text data from tweets. The multimodal nature of this approach provides better results in the presence of noise and it is able to perform recognition in the absence of image. The high performance of the model makes it perfect fit for the task at hand.

\subsection{Graph update rules}
\label{sec:graph_update_rules}
Updating the graph makes it larger and therefore, the memory graph approach is considered to shrink it to the desired size. However, adding multiple tweets (subgraphs) to it needs precautions, because each addition changes the shape and form of connections. Tracking changes and inferring topics according to small changes will be harder as the data stream get faster or the topic gets hotter. In order to detect and track topics, we propose using graph update rules that we obtained by getting motivated from \cite{zhao2019incremental}. They initially used this approach for community detection and we employ a different but similar method for topic detection.
Each tweet converted to a complete subgraph, is part of a larger graph called memory graph. At each time step $t$, incoming tweets are converted to complete subgraphs such as $G_t=(V_t,E_t)$. For simplicity, we see tweets coming in a specific time stamp one by one. Time stamp $t$ is the identifier for each complete subgraph such as $G_t$ generated from tweet. The $V_t$ symbol presents words or tokens and $E_t$ presents the connection between them. $E_t$ is complete for any tweet and is always determined according to \eqref{eq:score} and \eqref{eq:embedding}. Rest of the notations are explained in Table \ref{tb:notations}.

\begin{table*}[h]
\centering
\caption{Notations used in this paper}
        \label{tb:notations}
		\begin{tabular}{l l}
        \hline
        Notation & Description\\
        \hline
        $G_t=(V_t,E_t)$&Complete graph from twitter stream at time $t$ with tokens as $V$ and edges as $E$\\
        $S_{i,j}^t$&Cumulative similarity score between tokens $i$ and $j$ in time $t$ (Eq. \ref{eq:score})\\
        $M_i^t$&Embedding of token $i$ at time $t$ (Eq. \ref{eq:embedding})\\
        $\theta^t$&Set of all topic clusters at time $t$\\
        $\theta_i^t$&$i$th topic cluster at time $t$\\
        $\theta$&Set of all topic clusters in all times\\
        $\theta_i^t$&$i$th topic at time $t$\\
        $N^t(W)$&Neighbourhood of $W$ in time $t$\\
        \hline
        \end{tabular}
        
 \end{table*}
 
According to the assumptions above, we redefine each edge as a score in \eqref{eq:score} and note the subgraph as $G_t=(V_t,S_t)$ in which $S_t \equiv S_{i,j}^t$. To understand the graph update rules, we first categorize the incoming subgraph type and according to its type, we decide which strategy to apply. Table \ref{tb:categorization} described each category. The first category $\mathcal{U}$ needs no precaution for addition because it will form a completely new topic cluster $\theta_i^t \not \in \theta^t$ independent of any other previous one. The strategy required for this category is to assign a new cluster id for it and add it to the set of topic clusters at time $t$ that is noted as $\theta^t$. In other sense, $(\forall a) W_a \in V_t\rightarrow (\forall i) W_a \notin \theta_i^t$ or $(\forall i) V_t \cap \theta_i^t = \emptyset$ is the mathematical definition of this category. In the case of category $\mathcal{I}$, the new graph will be completely added to an existing $\theta_i^t$ because $(\exists i) V_t \subset \theta_i^t$. This update will change the inner structure of $\theta_i^t$ and has no impact on any other topic clusters. The Multiple category or $\mathcal{M}$ is identified as words belonging to more than one topic cluster. Its mathematical definition is expressed as $[(\forall a) W_a \in V_t\rightarrow (\exists i) W_a \in \theta_i^t]\wedge [(\forall j) V_t \not\subset \theta_j^t]$. The last category, $\mathcal{S}$, is expressed as $[(\exists a) W_a \in V_t\rightarrow (\forall i) W_a \notin \theta_i^t]\wedge[(\exists b) W_b \in V_t\rightarrow (\exists i) W_b \in \theta_i^t] \wedge [(\forall j) V_t \not\subset \theta_j^t]$. For the last two cases of these categorization, $\mathcal{M}$ and $\mathcal{S}$, we need algorithmic approach to determine which word belongs to which topic cluster that is presented by Algorithm \ref{alg:algall}. In this algorithm, the topic cluster stickiness metric is used that is explained in \eqref{eq:stick}. The $\Lambda$ function provides a metric to determine how much a single word $W_a$ is related to a topic such as $\theta_i^t$. This function can be used in different approaches, utilization of this function is described by Algorithm \ref{alg:algall}.

\begin{equation}\label{eq:stick}
\Lambda(W_a,\theta_i^t) = \frac{\sum\limits_{W_b \in N^t(W_a) \cap V_{\theta_i^t}}S_{W_a,W_b}^t}{\sum\limits_{W_b \in N^t(W_a)}S_{W_a,W_b}^t}
\end{equation}
\begin{algorithm}\label{alg:algall}
   \caption{Update $\theta^t$ for all the categories listed in Table \ref{tb:categorization}}
   \SetAlgoLined
   \LinesNumbered
  \SetKwInOut{Input}{Inputs}
  \SetKwInOut{Output}{Output}
  \SetKwProg{Update}{UpdateTopicClusters}{}{}
  \Update{$(\theta^t,G_t)$}{
    \Input{$\theta^t$ and $G_t=(V_t,S_t)$}
    \Output{$\theta^{t+1}$}
    $G_{unseen}=\{W \in V_t | (\forall i) W \notin \theta_i^t\}$\;
    $G_{seen}=V_t - G_{unseen}$\;
    $S_{update} =\{S_{W,W'}^t|W \in V_t$ and $W' \in N(W)\}$\;
    Apply $S_{update}$to related edges in update\;
    \If{$G_{seen} == \emptyset$}{
    $\theta_{K^t+1}^t=G_t$\;
    }
    \Else{
    $\theta_{K^t+1}^t=G_t$\;
    \ForEach{$\mathcal{G} \in [G_{unseen},G_{seen}]$}
    {
    \ForEach{ $W \in \mathcal{G}$}{
    assign $W$ to $\theta_i^t$ where $i= \argmax\limits_{i\in\{1,2,\dots,K^t+1\}} \{\Lambda(W,\theta_i^t)|N(W)\cap\theta_i^t \neq \emptyset\}$\;
    }
    }
    \If{$\theta_{K^t+1}^t ==\emptyset$}{
    remove $\theta_{K^t+1}^t$ from $\theta^t$\;
    }
    }
    \KwRet{$\theta^t$}\;
  }
\end{algorithm}

This algorithm provides a combination of all categorizations in a better understandable and efficient way. At first, if the incoming $G_t$, consists of all new words, if it is type $\mathcal{U}$, a new topic cluster is formed and assigned to it. Otherwise, for seen and unseen words in the $G_t$ (first seen and then unseen ones), we calculate $\Lambda$ for each $W$ and assign $W$ to the topic cluster which has the higher $\Lambda$. This part of algorithm is required for the $\mathcal{M}$ and $\mathcal{S}$ categories while for the $\mathcal{I}$ no update is required, just updating the $S$ values is enough. The reason for no required update in category $\mathcal{I}$ is that no new $W$ will be added to topic clusters, just the $S$ values will be updated.
\begin{table*}[h]
\centering
\caption{Categorization of incoming tweets as subgraphs}
        \label{tb:categorization}
		\begin{tabular}{l c l}
        \hline
        Category & Symbol & Description\\
        \hline
        Unique &$\mathcal{U}$& \textit{All} words in tweet are \textbf{new} and did not appear in any $\theta_i^t$\\
       Incessant &$\mathcal{I}$& \textit{All} words in tweet are \textbf{previously merged into a \underline{single $\theta_i^t$}}\\
        Multiple &$\mathcal{M}$& \textit{All} words in tweet are \textbf{previously merged into \underline{more than one $\theta_i^t$}}\\
        Subset &$\mathcal{S}$& \textit{Some} words in tweet are \textbf{previously merged into \underline{one or more that one $\theta_i^t$}} and \textit{some} are \textbf{new}\\
        \hline
        \end{tabular}
 \end{table*}

\subsection{Topic Extraction\label{sec:topic_extraction}}
The extraction part seems much easier having well-separated topic clusters which only contain related and connected words. However, using other features of the tweets is also helpful. By other features, we use how much tweet has been retweeted and liked. The other features are also available too, such as word frequency itself and the trend. We use all these features and store them in each node along with the embedding. Tagging words and grouping them as entities also improves the outcome. Here, we describe the methodology behind extracting topics from topic clusters at time $t$ or the $\theta^t$.
Equations \eqref{eq:topic_score} and \eqref{eq:node_score} describe how we score each node in each time $t$.

\begin{equation}\label{eq:topic_score}
\Gamma^t(W) = E\times\left(\Upsilon^t(W) + \sum\limits_{W'\in N^t(W)}\Upsilon^t(W')\times S^t_{W,W'}\right)
\end{equation}
\begin{equation}\label{eq:node_score}
\Upsilon^t(W)=\delta F^t\times\left(F^t(W) +\sum\limits_{i=0}^t\left(L_i+R_i\right)\right)
\end{equation}\begin{equation}
E=
\begin{cases}
1.2, & \text{if}\ W \text{ is tagged as named entity} \\
1, & \text{otherwise}
\end{cases}
\end{equation}

From the above equations, the outcome values for $\Gamma$ are converted to probabilities by \eqref{eq:p_w}. In this equation, $\theta^t_i$ is the topic cluster that $W$ belongs to. For each cluster, the probability of that topic cluster is in the topics is computed by \eqref{eq:prob_topic} and also, for each topic cluster, the probability of each word being nominated is computed by \eqref{eq:prob_word}. Multiplication of these two values gives the probability of each word being nominated in the overall outcome. Higher ranked words will appear first.

\begin{equation}\label{eq:p_w}
P(W|\theta^t_i) = \frac{\Gamma^t(W)}{\sum\limits_{W'\in \theta^t_i}\Gamma^t(W')}
\end{equation}
\begin{equation}\label{eq:prob_topic}
P(\theta^t_i|\theta^t) = \frac{\sum\limits_{W\in \theta^t_i}\Gamma^t(W)}{\sum\limits_{W\in \theta^t}\Gamma^t(W)}
\end{equation}
\begin{equation}\label{eq:prob_word}
P(W|\theta^t) = P(W|\theta^t_i)\times P(\theta^t_i|\theta^t)
\end{equation}

\subsection{Proposed system in a bounded harmony}
Although all parts and sections describe the system in little pieces but how this parts are bounded needs to be clarified. Figure \ref{fig:proposed_system} shows a visualization of the proposed system. The most important part of this systems is the data flow into the parts and keep it in a harmony. By harmony we mean how each part does its task and minimize the requirement for delays caused by slower parts.

First, the connection to the twitter data stream is established and the incoming tweets about a topic of interest (this can be done by targeting specified keywords or just querying on specific key-phrases) are stored in the MongoDB as they are collected. These batches of data contain very valuable fields such text, media, hyperlinks and etc. which we just select the necessary fields and send them to each module that is indicated on the connection curves in the figure. Each batch containing 64 text samples are passed into BERT for fine-tuning. The masked language modeling using this transformer fine-tunes the embeddings in our use case. On the other hand, the graph strategies defined in Section \ref{sec:graph_update_rules} is used and applied by Algorithm \ref{alg:algall}. The named entity recognition is done by using both image and text from this multimodal data stream using what is described in Section \ref{sec:multimodalner}. Graph strategies module adds data into the graph with doing topic clustering at the same time while the NER tags related words and combines them if there is an entity containing more than one words. We used Neo4j as our database because of its similicity in use and the ability to keep larger amounts of data while helping to access them using Cipher query language. This database is where we collect our findings in a graph form and share it between modules. The mining/tuning module is described in Section \ref{sec:topic_extraction} which after mining the topics using equation \eqref{eq:prob_word}, applies the decay defined by \eqref{eq:forgettingcurve} to nodes. As hyperparameter for forgetting, we used a dynamic method that investigates how much RAM is available and according to that finds new $\varrho$. If the memory usage goes over the threshold ($90\%$) we start reducing the $\varrho$ by 1 until we get to the exact RAM usage after automatic elimination of nodes. If it goes under the threshold ($80\%$) the $\varrho$ is increased by 1 until it is reached to the desired value.The initial value of $\varrho$ is set to $10^3$. The results are sent to the output at each step mining/tuning has been applied.

This independence of modules makes them work concurrently or on different servers if needed.
\section{Experiments and results}
\label{sec:experimentalResults}
For experimenting our system on various datasets we have used the datasets provided by \cite{aiello2013sensing} which was part of \textit{social sensor project}. The three collected datasets along with results all of the methods that have been proposed upon these datasets are discussed in the following.
\subsection{Datasets} \label{sec:datasets}
According to the TREC practice, releasing contents of tweets are not allowed because it violates the user privacy. The user might change his preferences to not seen by public in future time and releasing in past will violate his/her rights eventually. Therefore, the authors did not release the tweets, but instead they released the tweet id to be accessed if it is still available. However, in time, small part of this dataset was not accessible because of many reasons such as user changes to the privacy settings, user account closed or even restricted. This issue is not specific to this dataset and happens when the providers of datasets follow the TREC rules and respect user privacy. At the time we collected these datasets, the same issue happened and the quantitative results about the dataset itself is reported in table \ref{tab:datasets}.

\begin{table*}[h]
\centering
\caption{Quantitative information about the datasets, duration time format is HH:MM:SS}
\label{tab:datasets}
\begin{tabular}{l l r l l l}
\hline
Dataset & Total Topics & Number of Tweets&Start Data&End Date&Duration\\
\hline
FA CUP &13&189,034&2012/05/05 14:00:00&2012/05/05 19:59:30&05:59:30\\
SUPER TUESDAY&22&475,659&2012/03/06 17:00:00&2012/03/07 16:59:59&24:00:00\\
US ELECTIONS&64&1,992,260&2012/11/06 17:00:00&2012/11/08 04:59:59&36:00:00\\
\hline
\end{tabular}
\end{table*}

\begin{figure*}[!ht]
	\centering
	\includegraphics[width=0.25\linewidth]{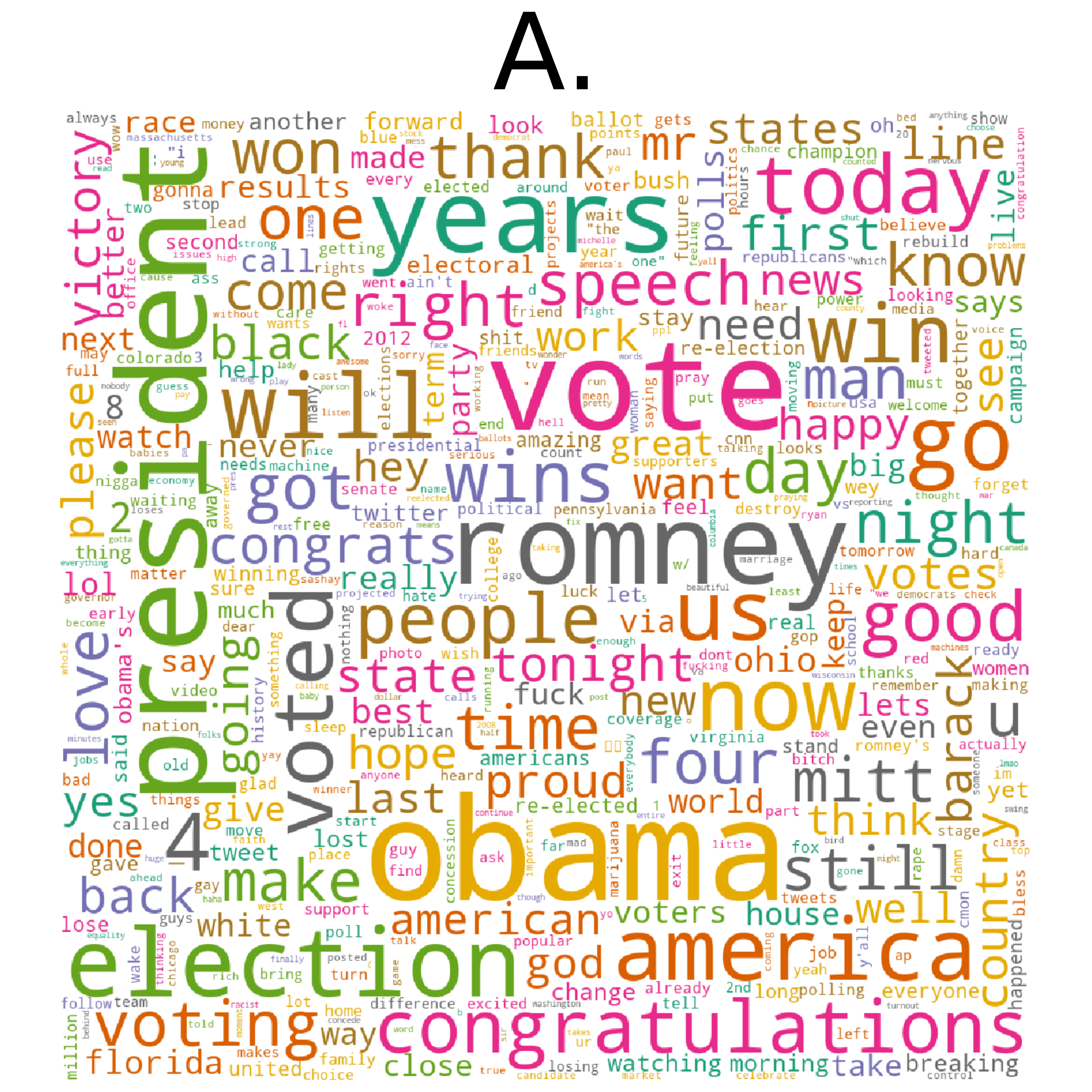}
	\includegraphics[width=0.25\linewidth]{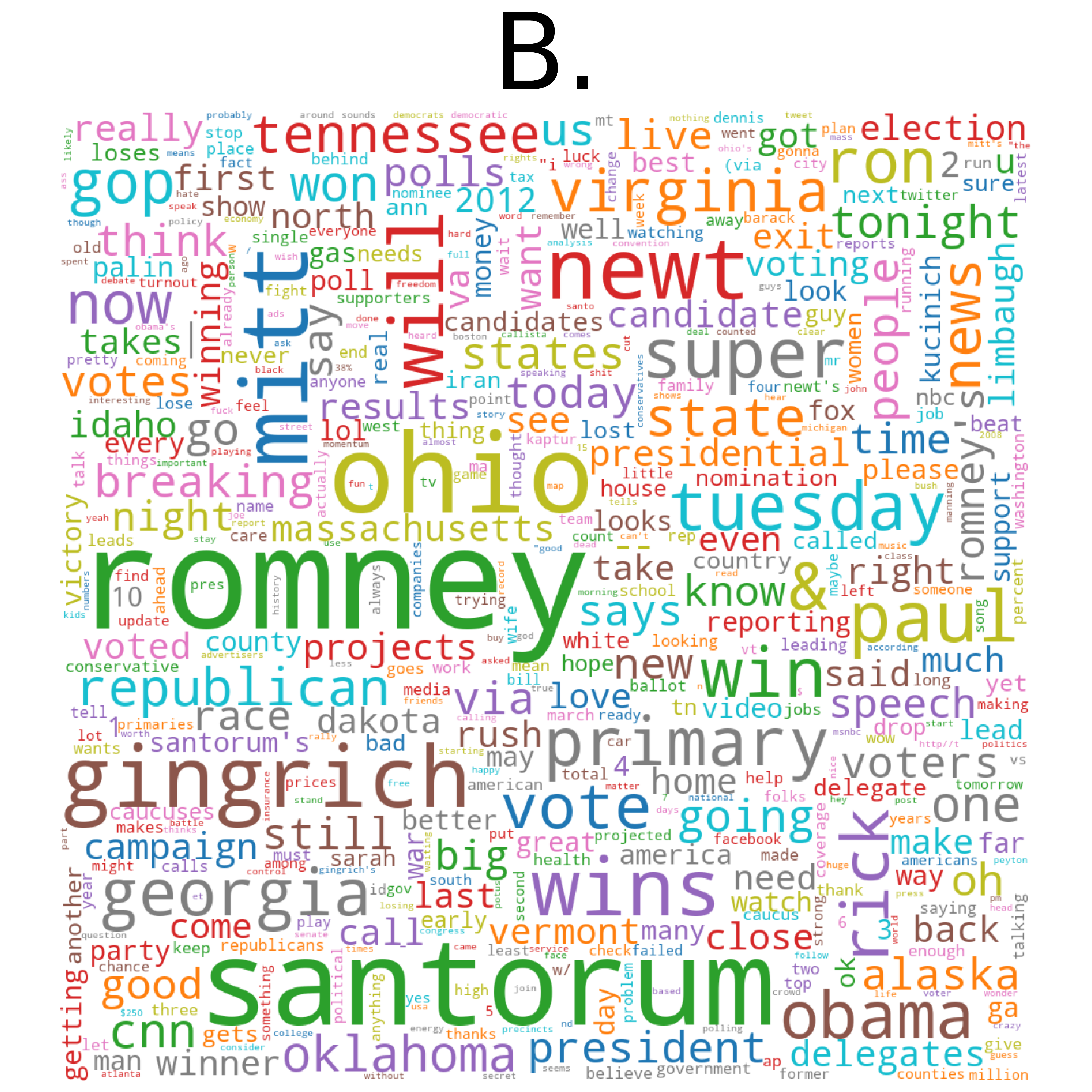}
	\includegraphics[width=0.25\linewidth]{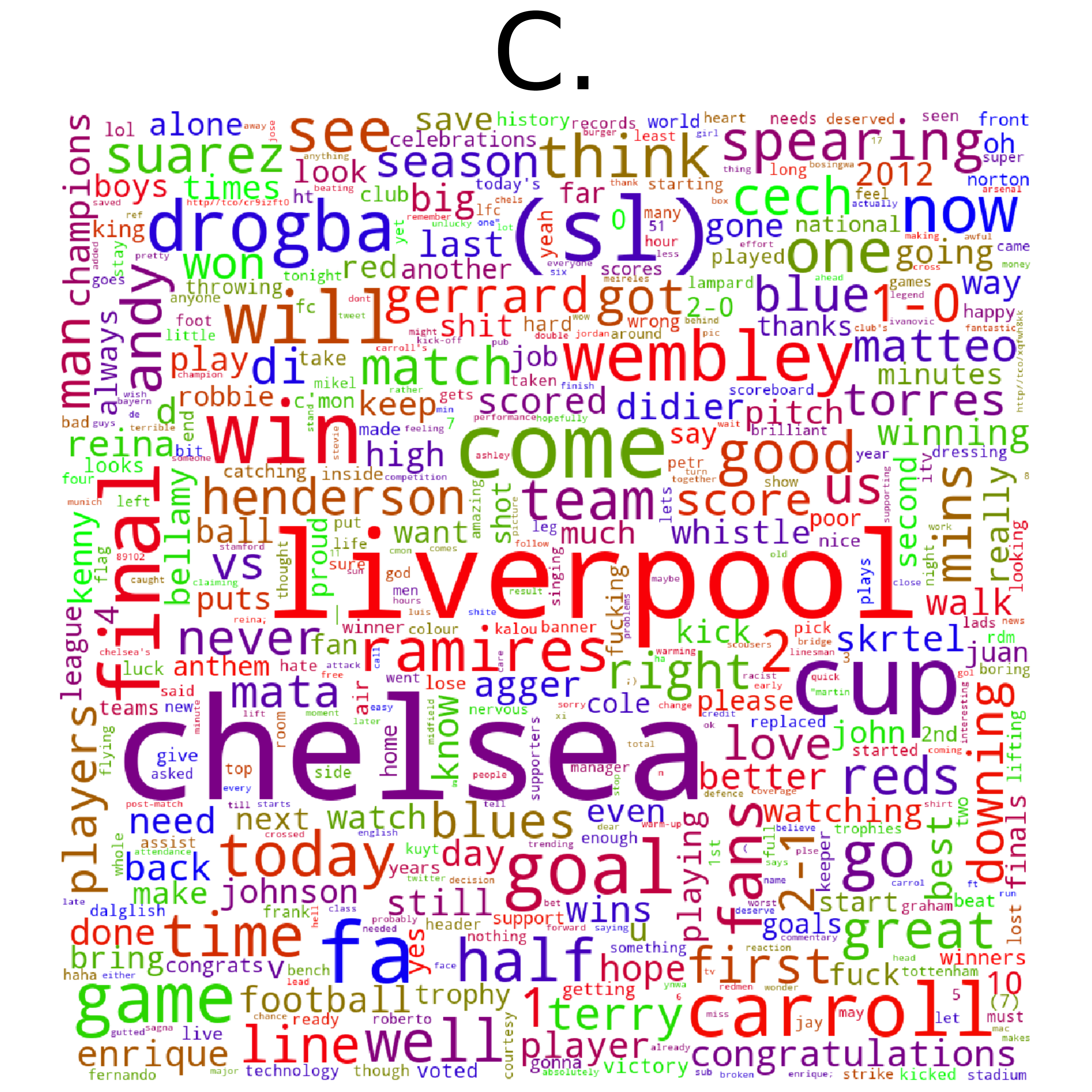}
	\caption{WordCloud representation of datasets: A. US Elections; B. Super Tuesday; C. FA Cup.}
	\label{fig:wordcloud_datasets}
\end{figure*}

\begin{figure*}[!ht]
	\centering
	\includegraphics[width=0.25\linewidth]{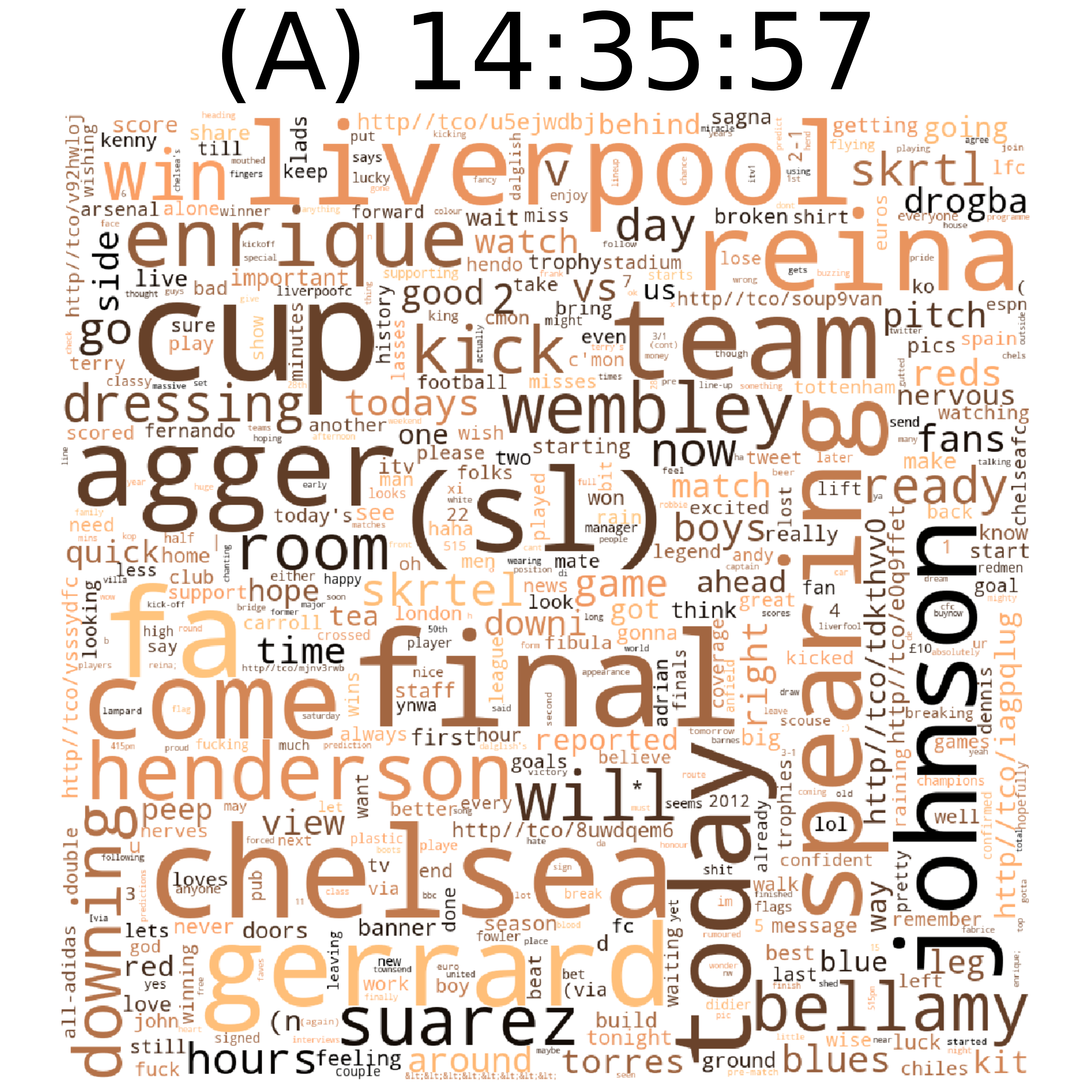}
	\includegraphics[width=0.25\linewidth]{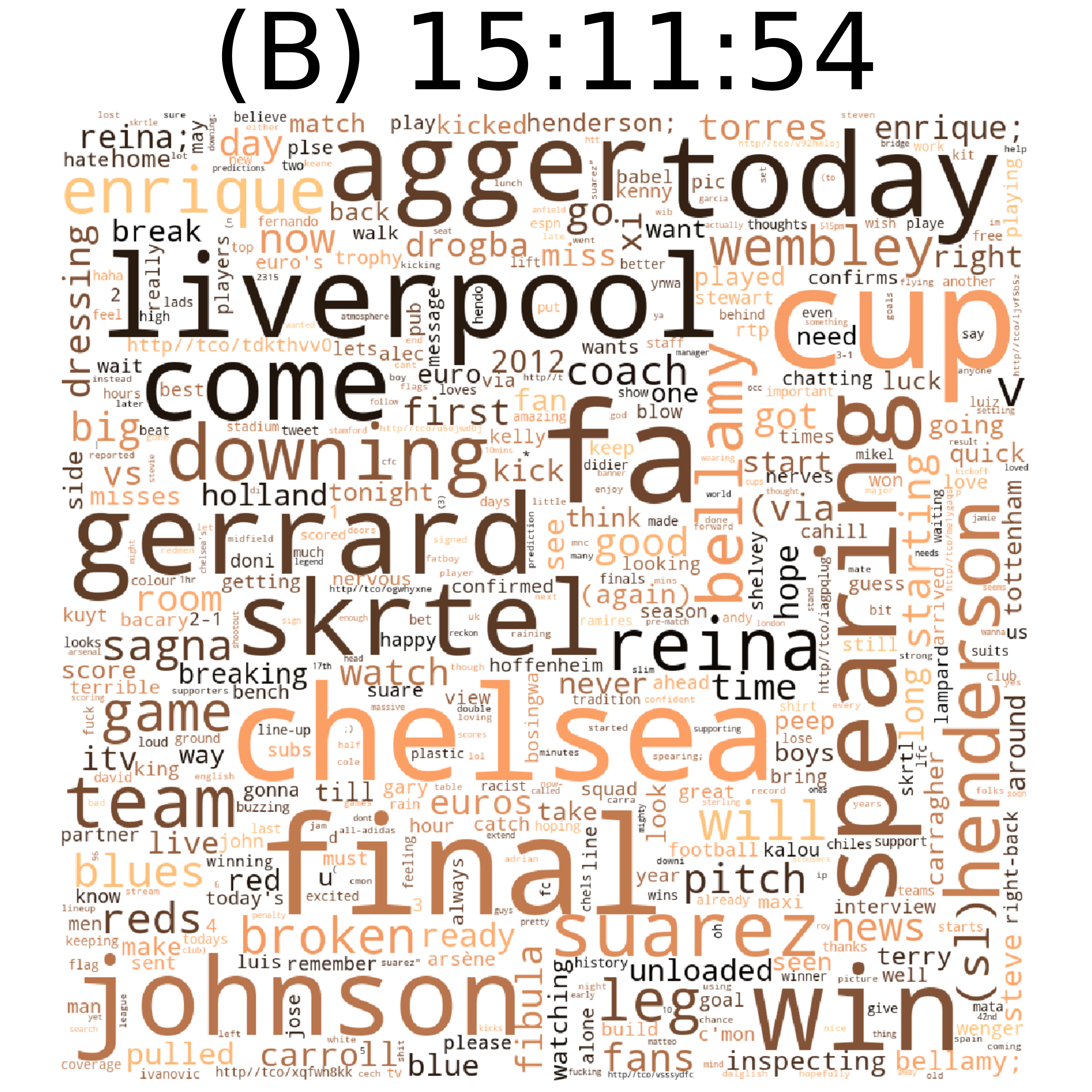}
	\includegraphics[width=0.25\linewidth]{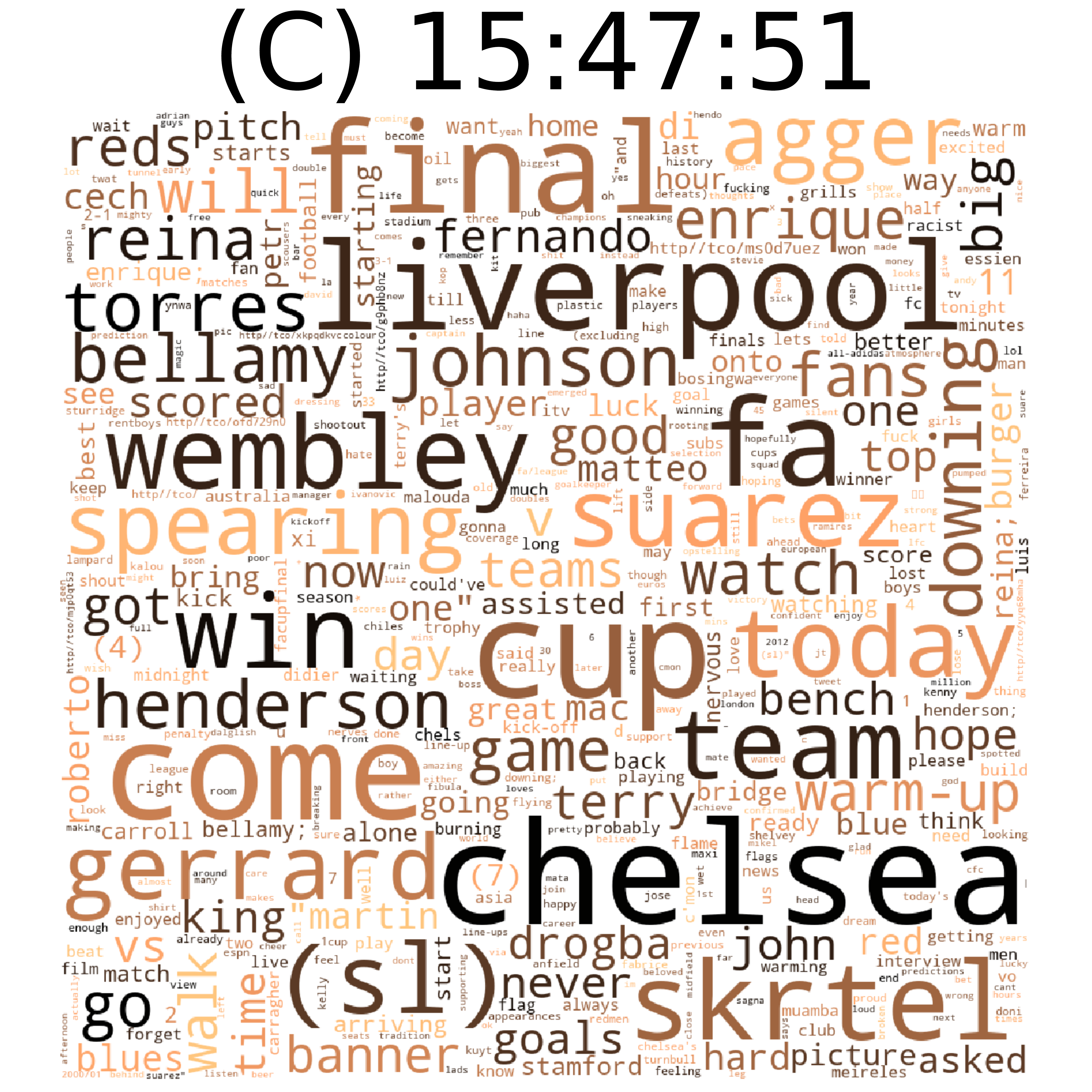}
\\
	\includegraphics[width=0.25\linewidth]{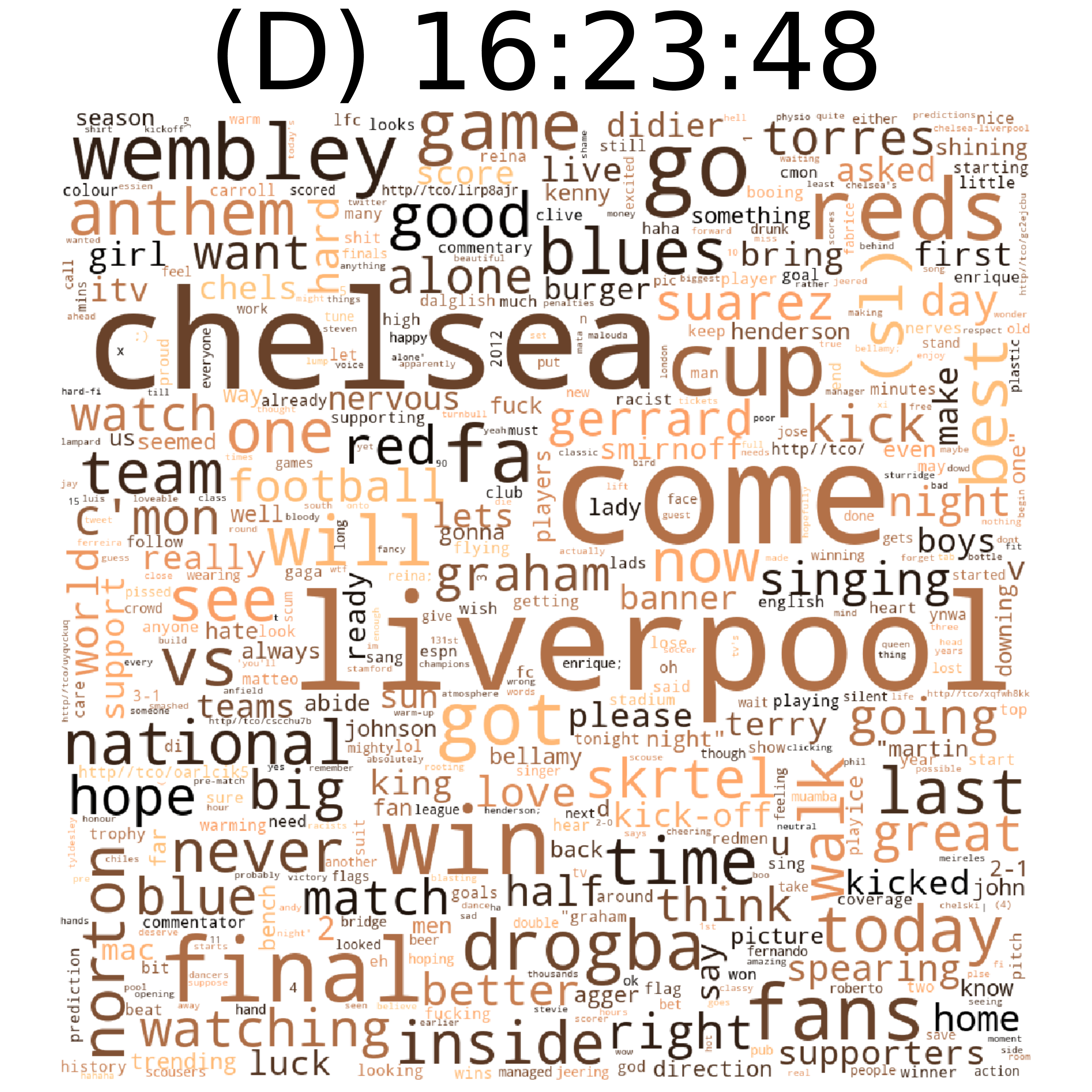}
	\includegraphics[width=0.25\linewidth]{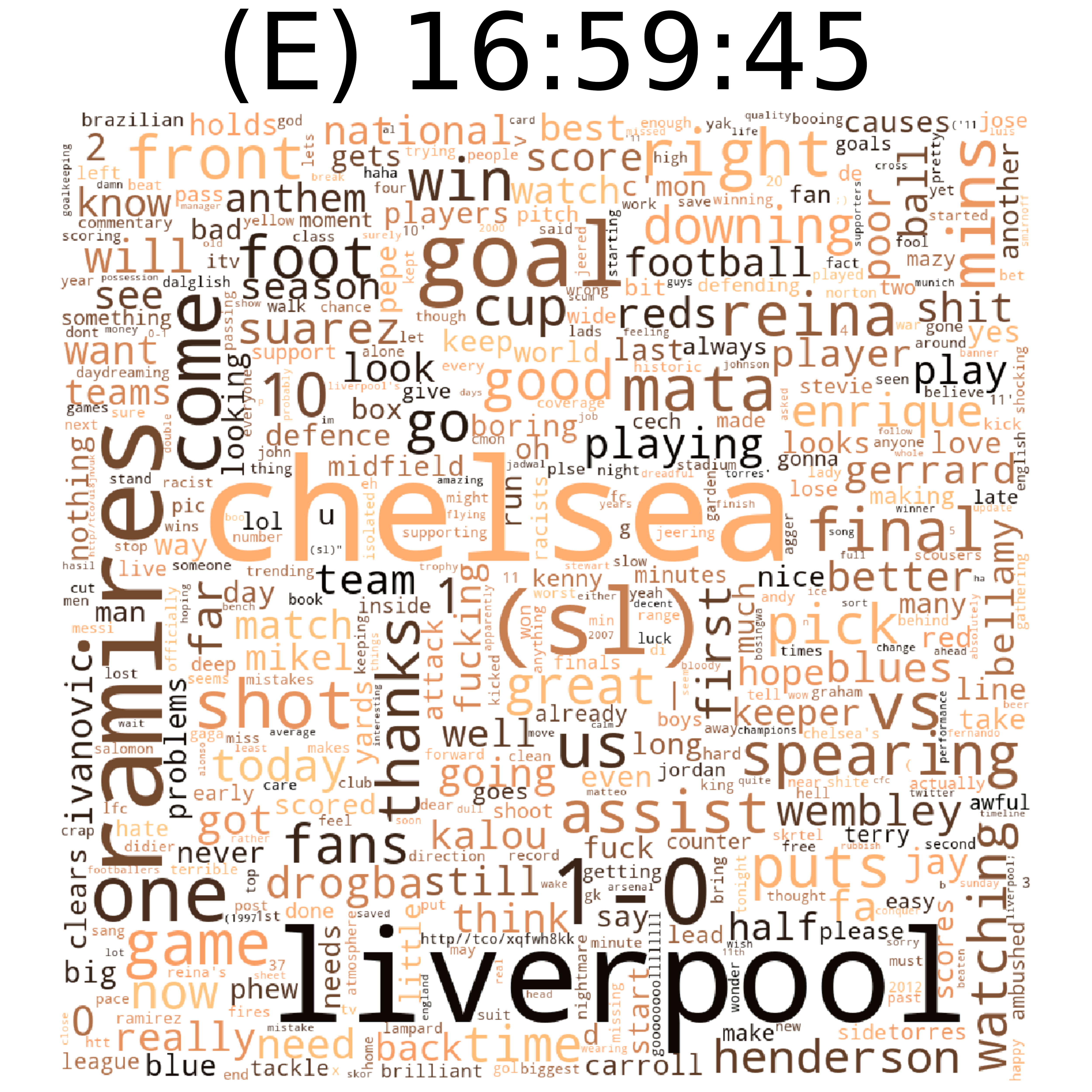}
	\includegraphics[width=0.25\linewidth]{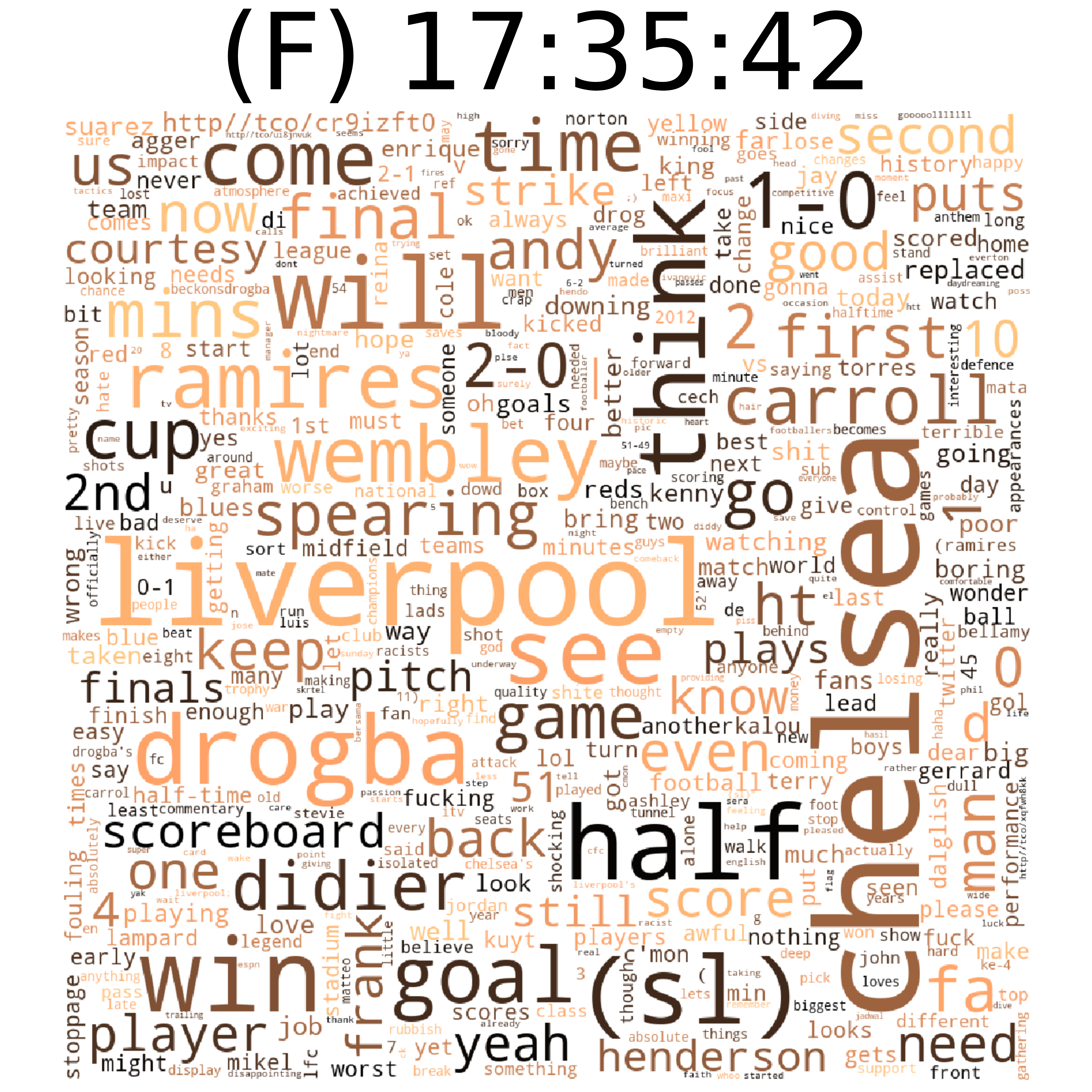}
\\
	\includegraphics[width=0.25\linewidth]{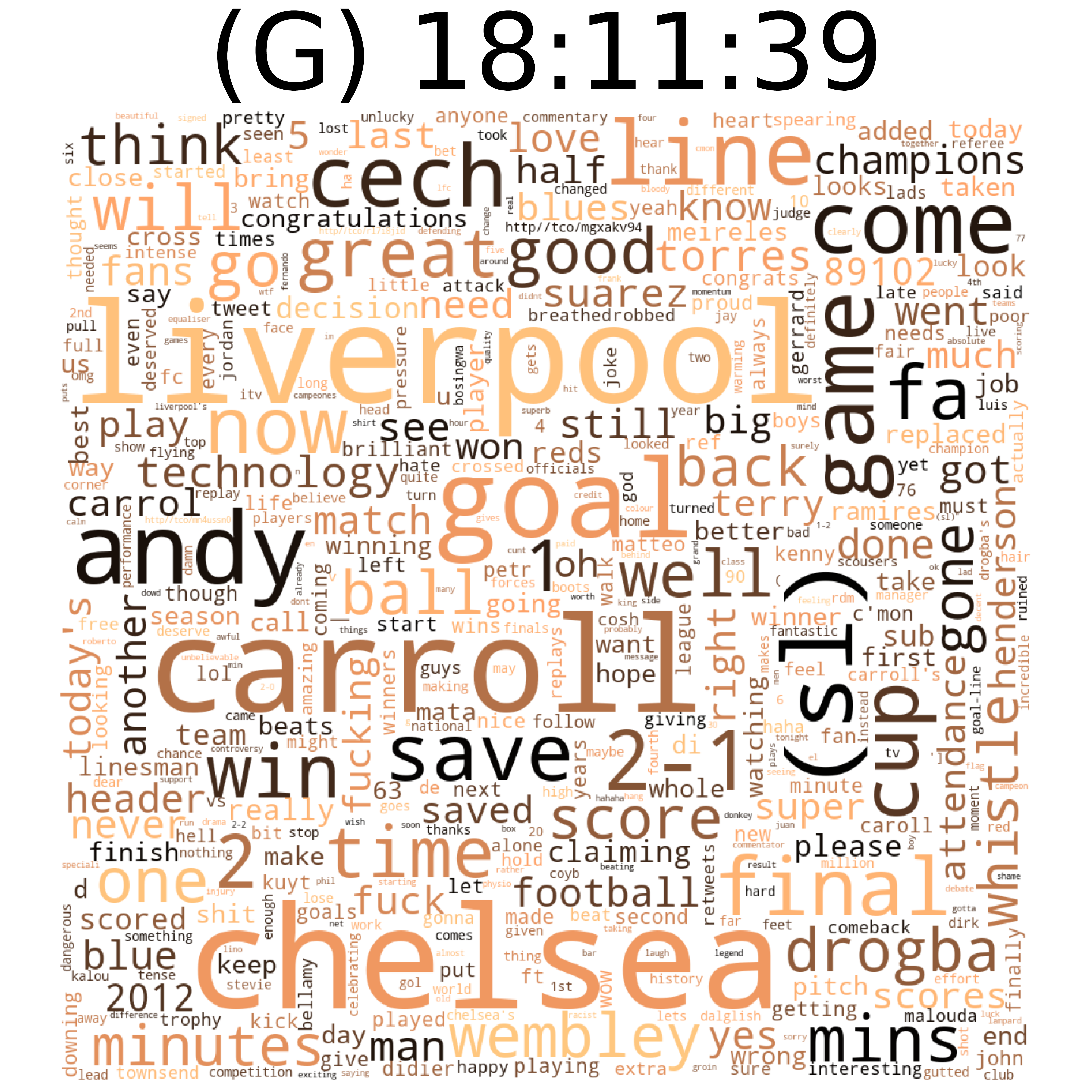}
	\includegraphics[width=0.25\linewidth]{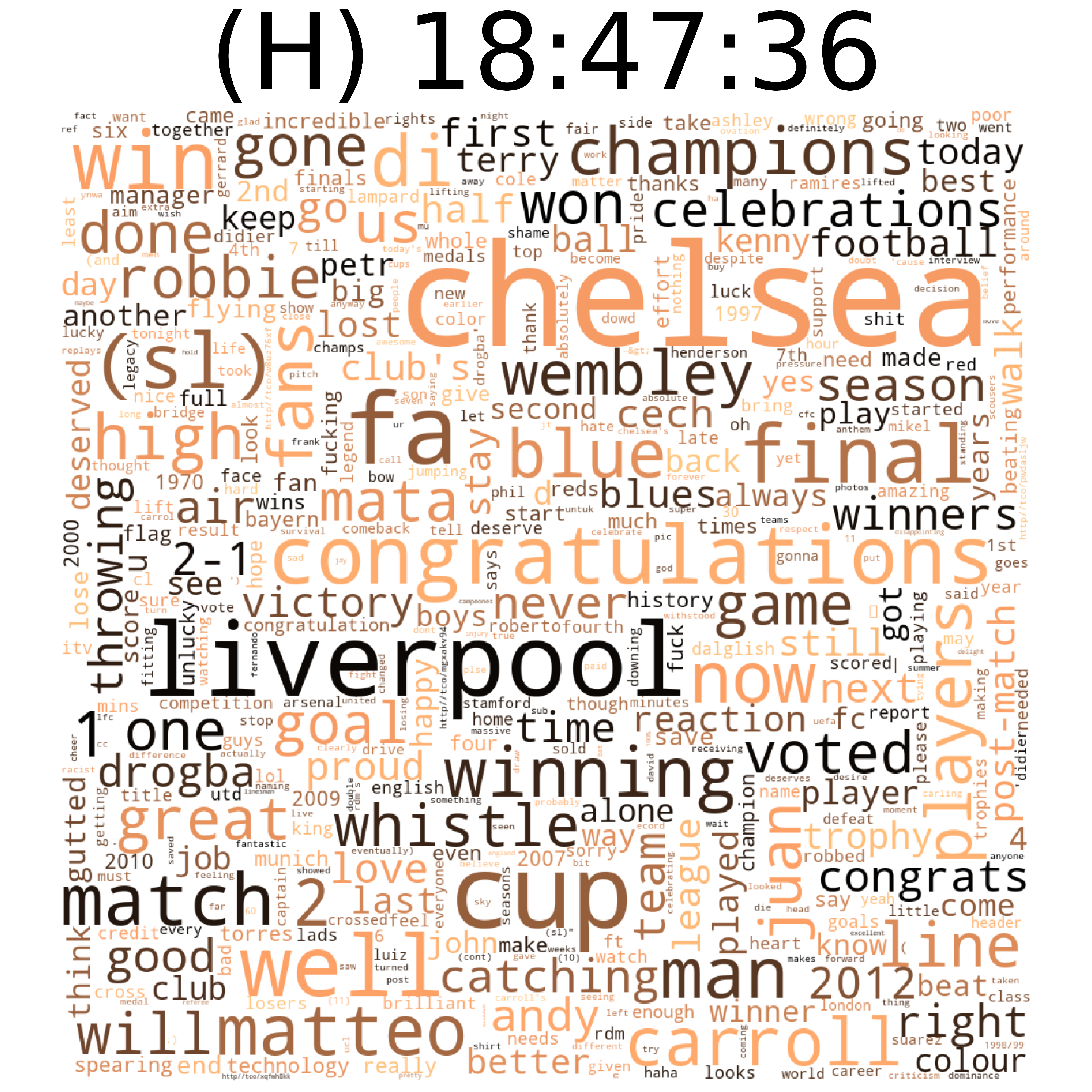}
	\includegraphics[width=0.25\linewidth]{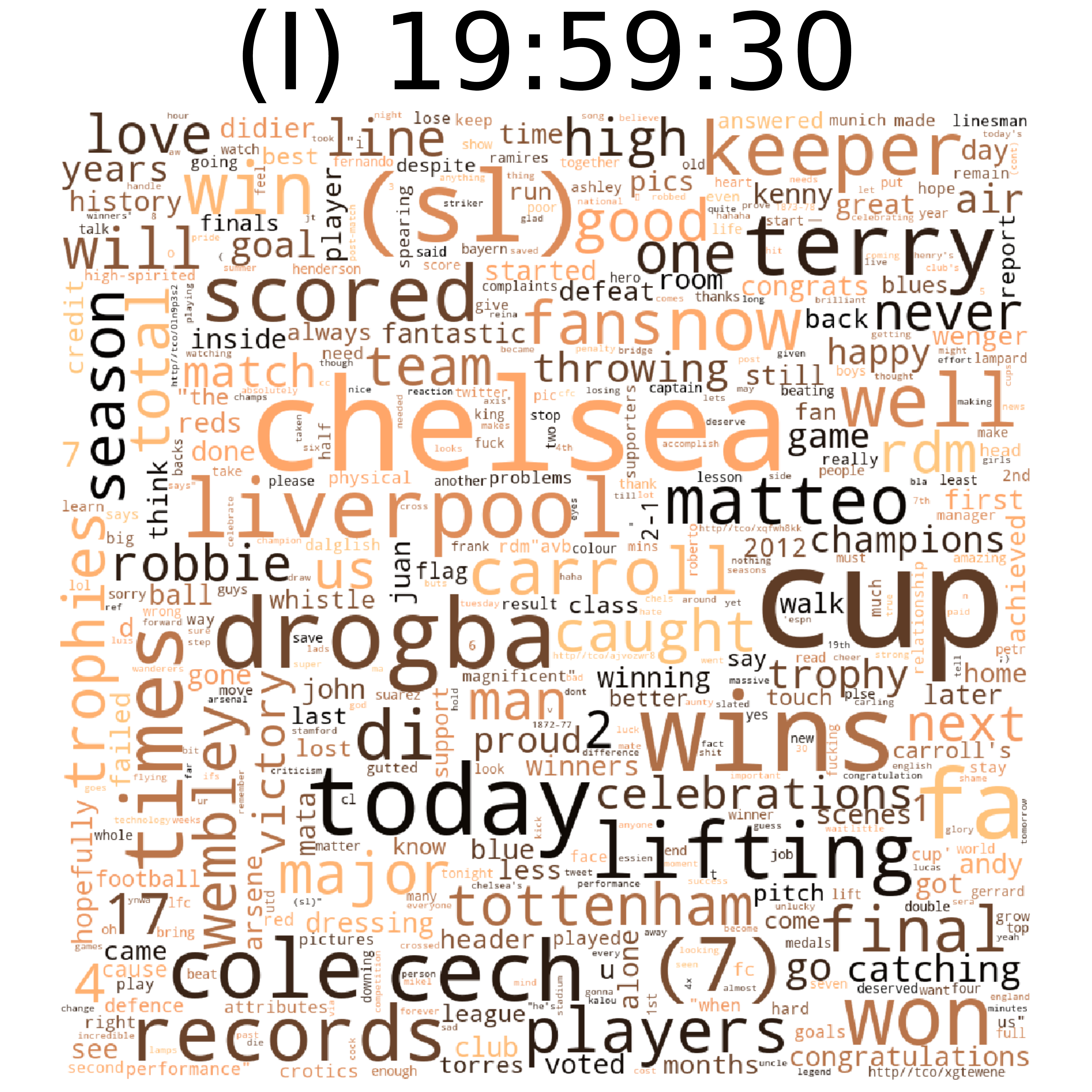}
	\caption{WordCloud representation of FA Cup dataset}
	\label{fig:wordcloud_fa_cup}
\end{figure*}

\subsection{Preprocessing}
Tweets often contain huge amounts of noise such as mistyped and user invetned words. This noise is a big problem in cases where no subword embedding is used. One of reasons we used BERT is because it can handle such unknown tokens. For the preprocessing part, we dropped stop words if they do not appear in the named entities. Other than dropping stopwords, we also dropped non-english tweets that contain various languages such as Spanish or Hindi and etc. We considered such languages out of our scope but our method is capable of handling these methods and we put into our future works to use multilingual contextual pretrained models\footnote{Pretrained multilingual BERT available at: \url{https://github.com/google-research/bert/blob/master/multilingual.md}}. Cleaning other characters such as \texttt{@} or \texttt{-} did not damage results more than 0.05 in various metrics but dropping these caharacter from tokens would yield in losing important tokens such as 1-0 in datasets such as FA CUP. Because of event itself, designing such preprocessing and engineering it is a hard thing to do and in some cases (rather than datasets we have tested and in some other languages) it may be beneficial but in overall it is logical to keep it as minimum as possible.

\begin{table*}
\centering
\caption{Top-K topics-recall of detection algorithms compared to our proposed method on FA CUP dataset}
\label{tb:res_facup}
\begin{tabular}{l l l l l l l l l l l}
\hline
\multirow{2}{*}{Method}&\multicolumn{10}{c}{Top-K topic-recall}\\
\cline{2-11}
&2&4&6&8&10&12&14&16&18&20 \\
\hline
LDA&.692&.692&.840&.840&.920&.920&.840&.840&.840&.750\\
Doc-P&.769&.850&.920&.920&1&1&1&1&1&1\\
Gfeat-P&.000&.308&.308&.375&.375&.375&.375&.375&.375&.375\\
SFPM&.615&.840&.840&1&1&1&1&1&1&1\\
BNGram&.769&.920&.920&.920&.920&.920&.920&.920&.920&.920\\
SVD+Kmean&.482&.596&.710&.824&.938&.951&.951&.951&.951&.951\\
SNMF-Orig&.100&.177&.254&.331&.389&.389&.389&.389&.389&.389\\
SNMF-KL&.167&.334&.502&.670&.837&.837&.840&.850&.850&.924\\
Exempler&.810&.838&.886&.908&.916&.916&.916&.916&.916&.916\\
EHG&.379&.591&.727&.727&.864&.864&1&1&1&1\\
Ours&.810&.951&.951&1&1&1&1&1&1&1\\
\hline
\end{tabular}
\end{table*}

\begin{table*}
\centering
\caption{Top-K topics-recall of detection algorithms compared to our proposed method on SUPER  TUESDAY dataset}
\label{tb:res_super_tuesday}
\begin{tabular}{l l l l l l l l l l l l}
\hline
\multirow{2}{*}{Method}&\multicolumn{11}{c}{Top-K topic-recall}\\
\cline{2-12}
&2&10&20&30&40&50&60&70&80&90&100 \\
\hline
LDA&.000&.000&.000&.180&.130&.130&.180&.280&.280&.370&.227\\
Doc-P&.227&.227&.310&.400&.460&.500&.500&.500&.540&.680&.680\\
Gfeat-P&.046&.045&.085&.180&.227&.280&.280&.280&.280&.280&.280\\
SFPM&.182&.182&.270&.325&.325&.325&.325&.325&.325&.325&.325\\
BNGram&.500&.500&.540&.540&.540&.540&.540&.540&.540&.540&.540\\
SVD+Kmean&.192&.236&.400&.488&.547&.580&.626&.666&.666&.666&.666\\
SNMF-Orig&.000&.045&.100&.183&.277&.277&.277&.320&.320&.363&.453\\
SNMF-KL&.000&.100&.183&.183&.318&.410&.366&.410&.453&.363&.410\\
Exempler&.246&.463&.538&.572&.586&.597&.600&.617&.638&.638&.638\\
EHG&.163&.408&.466&.540&.628&.674&.674&.699&.699&.711&.711\\
Ours&.463&.381&.580&.628&.699&.699&.699&.711&722&.722&.722\\
\hline
\end{tabular}
\end{table*}

\begin{table*}
\centering
\caption{Top-K topics-recall of detection algorithms compared to our proposed method on US ELECTIONS dataset}
\label{tb:res_us_elections}
\begin{tabular}{l l l l l l l l l l l l}
\hline
\multirow{2}{*}{Method}&\multicolumn{11}{c}{Top-K topic-recall}\\
\cline{2-12}
&2&10&20&30&40&50&60&70&80&90&100 \\
\hline
LDA&.109&.109&.185&.245&.220&.280&.325&.500&.475&.430&.460\\
Doc-P&.234&.234&.415&.505&.560&.615&.615&.690&.690&.720&.740\\
Gfeat-P&.078&.078&.140&.180&.180&.180&.180&.180&.180&.180&.180\\
SFPM&.359&.359&.465&.525&.540&.540&.540&.540&.540&.540&.540\\
BNGram&.480&.480&.495&.495&.495&.495&.495&.495&.495&.495&.495\\
SVD+Kmean&.110&.216&.420&.522&.588&.608&.647&.700&.720&.720&.740\\
SNMF-Orig&.075&.075&.154&.218&.439&.467&.483&.545&.563&.595&.595\\
SNMF-KL&.154&.154&.326&.400&.547&.581&.562&.618&.600&.652&.622\\
Exempler&.022&.142&.244&.364&.465&.532&.590&.628&.651&.662&.662\\
EHG&.279&.608&.670&.688&.733&.746&.762&.772&.780&.796&.805\\
Ours&.326&.628&.628&.733&.772&.772&.780&.805&.805&.813&.821\\
\hline
\end{tabular}
\end{table*}

\begin{table*}
\centering
\caption{Top-2 keyword-precision of detection algorithms compared to our proposed method on all three datasets}
\label{tb:res_all}
\begin{tabular}{l l l l}
\hline
\multirow{2}{*}{Method}&\multicolumn{3}{c}{Dataset}\\
\cline{2-4}
&FA CUP&SUPER TUESDAY&US ELECTIONS\\
\hline
LDA&.164&.000&.165\\
Doc-P&.337&.511&.401\\
Gfeat-P&.000&.375&.375\\
SFPM&.233&.471&.241\\
BNGram&.299&.628&.405\\
SVD+Kmean&.242&.367&.300\\
SNMF-Orig&.330&.241&.241\\
SNMF-KL&.242&.164&.164\\
Exemplar&.300&.485&.391\\
EHG&.442&.812&.591\\
Ours&\textbf{.456}&\textbf{.851}&\textbf{.621}\\
\hline
\end{tabular}
\end{table*}

\subsection{Evaluation metrics and results}
\begin{figure*}[!ht]
	\centering
	\includegraphics[width=0.9\linewidth]{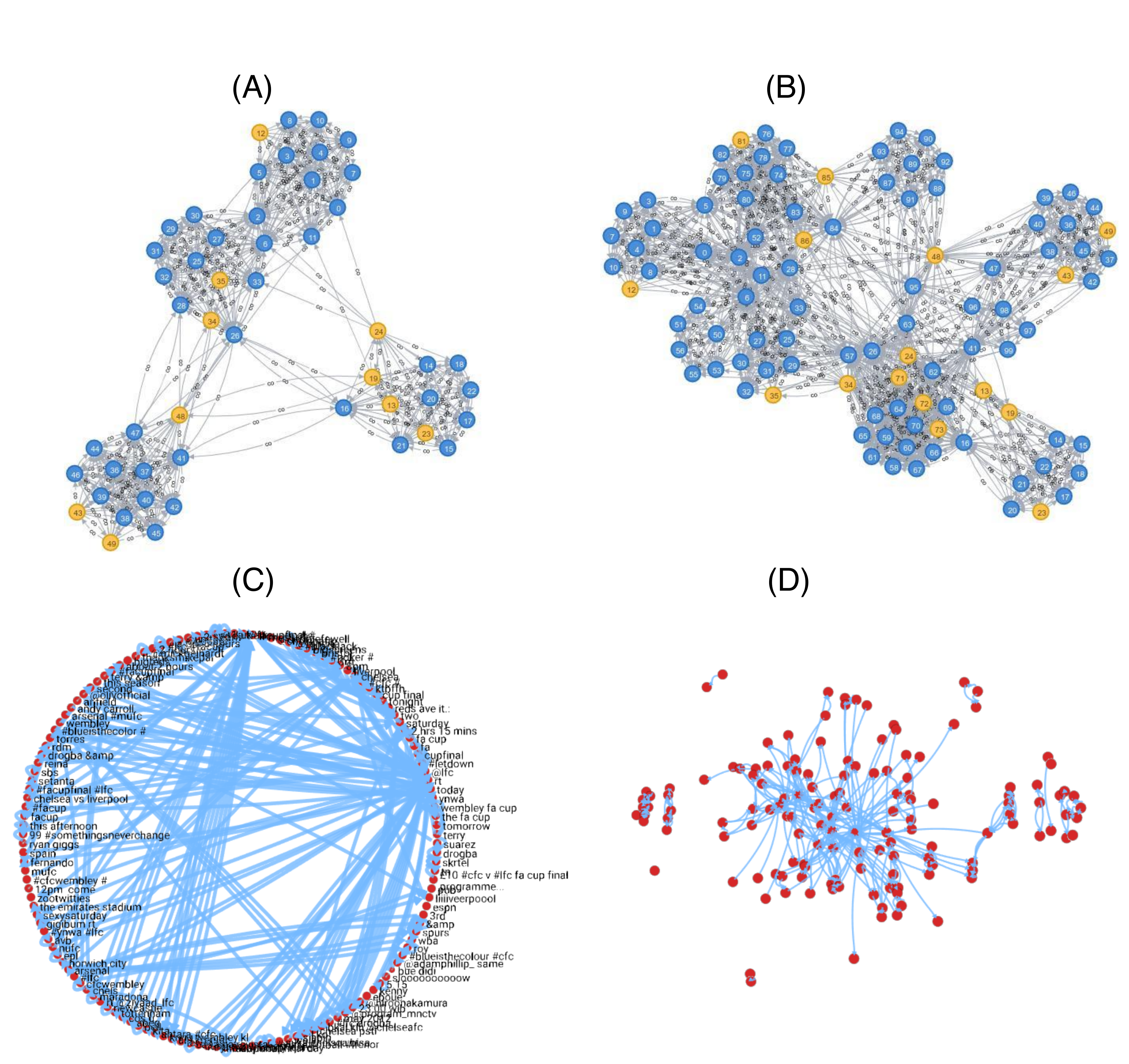}
	\caption{Outputs from Neo4j: (A) and (B) using Neo4j default visualization tool, yellow nodes are named entities and blues ones are words; (C) and (D) using GraphXR, all nodes are named entities}
	\label{fig:mg}
\end{figure*}

We use two metrics for evaluating our work, namely keyword-precision and topic-recall. Keyword-precision is defined as the number of correctly detected keywords divided by the number of ground-truth keywords. Topic-recall is the number of detected ground-truth topics divided by all ground truth topics. In order to evaluate and compare our work to others, we used the results obtained by other methods and also added our own on various top-K tests. For results of other methods, we used the reported values from \cite{saeed2019enhanced}. Tables \ref{tb:res_facup}, \ref{tb:res_super_tuesday}, and \ref{tb:res_us_elections} present results of our expriments on top-k topic-recall over all three datasets while table \ref{tb:res_all} provides results for all datasets with top-2 keyword-precision applied to these datasets. For the visualization part of our memory graph, we obtained results from our Neo4j database that is shown for an early moment before and after detecting communities. Figure \ref{fig:mg} shows this visualization, in  yellow nodes indicate the named entities while the blue ones are regular words.

\begin{figure*}[!ht]
	\centering
	\includegraphics[width=0.9\linewidth]{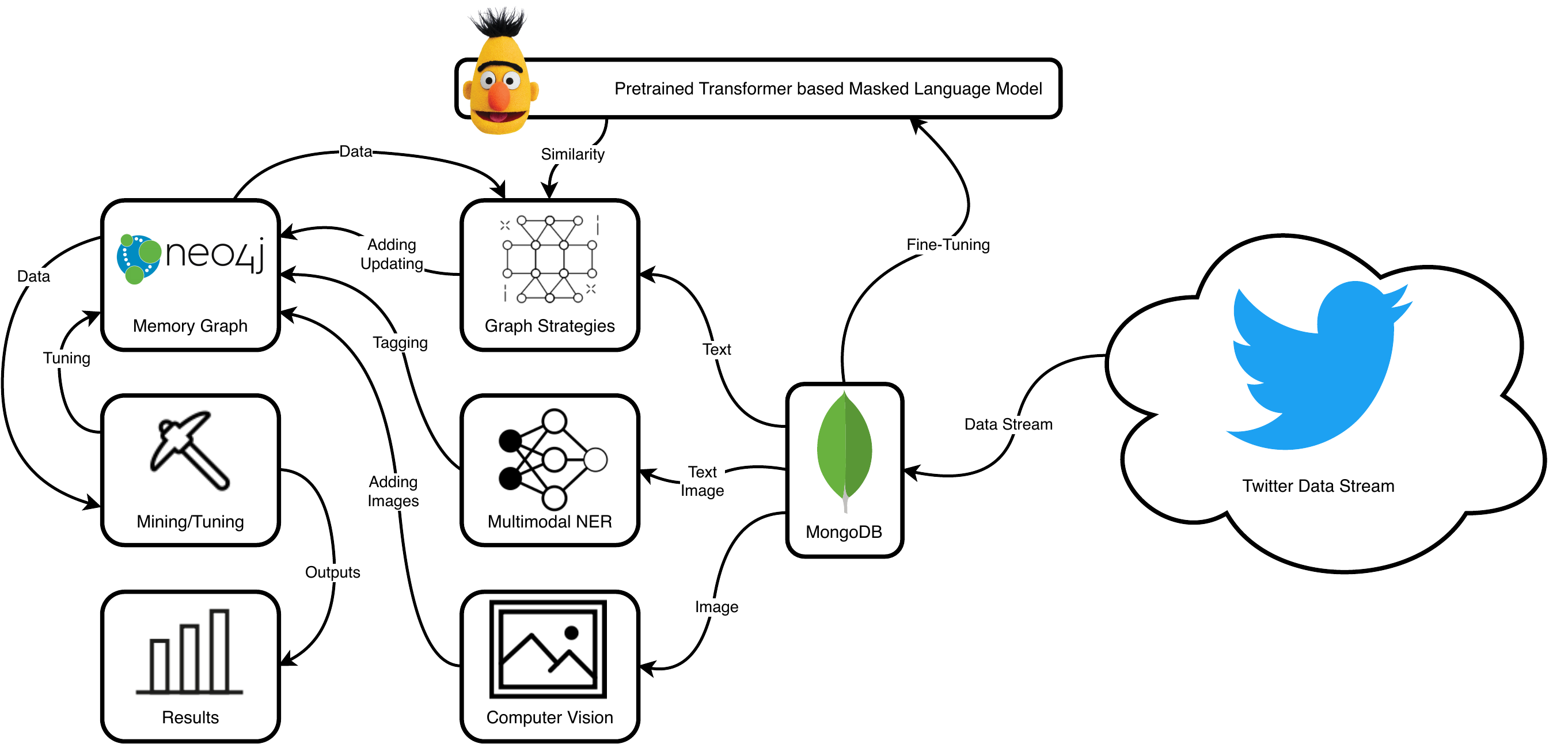}
	\caption{Proposed System}
	\label{fig:proposed_system}
\end{figure*}

\subsection{Future works}
In Figure \ref{fig:proposed_system}, we propose a module in our system as computer vision. This module is our future work guide for researcher to obtain other visual entities such as people and objects in visual content such as videos and images. Clustering this objects using their embedding would yield to having them bounded with the topics they appear most. Thus, such results would give the end user much more information rather than text and this visual content would help automatic journalism from social media.
\section{Conclusions}
\label{sec:conclusions}
In current research, we used Transformers combined with graph techniques to provide a topic detection system for social media. Fine-tuning the transformers in realtime and obtaining valuable semantic information rather than using just frequencies is another novelty of our work. Our proposed model is combined of various modules including BERT, graph strategies and multimodal named entity recognizer. This combination as a unique methodology is called memory graph that uses cognitive memorization in human brain. Using our hyper-parameters such as the rate of forgetting makes this system available for any stream size and any machine it is running. Modularity of our work makes it more usable in real life and corporate based environments when dealing with big social data.The results shows superiority of our proposed method compared to other state of the art techniques. 
	
	\begin{table*}
		\begin{tabular}{l p{3cm} c| c| c| c c p{3cm}}
			\hline
			\multirow{2}{*}{Ref.} & \multirow{2}{*}{Detection Method} & \multicolumn{2}{c}{Type} & \multicolumn{2}{c}{Task} & \multirow{2}{*}{Dataset} & \multirow{2}{*}{Application} \\
			\cline{3-6}
			&  & Event & Topic &  RED & NED &  & \\
			
			\hline
			\cite{sankaranarayanan2009twitterstand} & Naive Bayse Classifier & & \checkmark & & \checkmark & Twitter API, Handpicked Users & Hot News Detection \\
			\hline
			\cite{phuvipadawat2010breaking} & BScore based BOW Clustering  &\checkmark &  & & \checkmark & Twitter API (offline) & Disaster and Story Detection \\
			\hline
			\cite{petrovic2010streaming} & BOW Distance Similarity & \checkmark &  & & \checkmark & Twitter API & FSD \footnote{First Story Detection}  \\
			\hline
			\cite{tembhurnikar2015topic} & BN-Gram and tf-idf &  & \checkmark & \checkmark &  & Offline Datasets & Topic Detection \\
			\hline
			\cite{osborne2012bieber} & Cross Checking via Wikipedia & \checkmark &  &  & \checkmark & Twitter API, Wikipedia & Hot News Detection \\
			\hline
			\cite{cigarran2016step} & Formal Concept Analysis &  & \checkmark &  & \checkmark &  Replab 2013 dataset & Topic Detection \\
			\hline
			\cite{adedoyin2016rule} & FPM\footnote{Frequent Pattern Mining} & \checkmark &  &  & \checkmark &  Twitter API & Event Detection \\
			\hline
			
			\cite{Petkos2014} & FPM &  & \checkmark & \checkmark &  &   Super Tuesday/FA Cup/U.S. Elections & Topic Detection \\
			\hline
			
			\cite{dong2017discovering} & FPM (Hierarchical Clustering) &  & \checkmark &  & \checkmark &   topic dataset from CLEar system & Topic Detection \\
			\hline
			
			\cite{Gaglio2015} & FPM (tf-idf \& n-gram improved) & \checkmark &  &  & \checkmark &   Twitter API & Event Detection \\
			\hline
			
			\cite{erra2015approximate} & GPU improved tf-idf approximation &  & \checkmark & \checkmark &  &    offline dataset & Topic Detection \\
			\hline
			
			\cite{becker2011beyond} & BOW Similarity &  \checkmark &  &  & \checkmark &    offline dataset & Topic Detection \\
			\hline
			
			\cite{tang2014learning} & Word Embedding &   &  &  &  & SemEval Dataset & Twitter Sentiment Classification \\
			\hline
			
			\cite{hua2016automatic} &  spatiotemporal detection & \checkmark &  & \checkmark &  & offline Dataset & Targeted-domain event detection \\
			\hline
			
			\cite{abdelhaq2013eventweet} &  Clustering of Temporal \& Spatial features & \checkmark &  & \checkmark &  & Twitter API & Social Event Detection \\
			\hline
			
			\cite{lee2010measuring} &  Geographical Regularity Estimation & \checkmark &  &  & \checkmark & Twitter API & Geo-Social Event Detection \\
			\hline
			
			\cite{li2012tedas} &  BOW clustering & \checkmark &  &  & \checkmark & Twitter API & Event Detection \& Analysis \\
			\hline
			
			\cite{sakaki2010earthquake} &  Probabilistic Modeling & \checkmark &  &  & \checkmark & Twitter API & Early Disaster Detection \\
			\hline
			
			\cite{snowsill2010finding} &  FPM & \checkmark &  & \checkmark &  & offline dataset & Event Detection \\
			\hline
			
			\cite{cordeiro2012twitter} &  wavelet analysis & \checkmark &  &  & \checkmark & Twitter API & Event Detection \\
			\hline
			
		\end{tabular}
		\caption{Twitter Topic/Event detection/tracking related studies}
		\label{tab:tableRelatedWorks}
	\end{table*}

\bibliographystyle{elsarticle-num-names}

\bibliography{elsarticle-template-1-num}

\begin{thebibliography}{47}
\expandafter\ifx\csname natexlab\endcsname\relax\def\natexlab#1{#1}\fi
\providecommand{\url}[1]{\texttt{#1}}
\providecommand{\href}[2]{#2}
\providecommand{\path}[1]{#1}
\providecommand{\DOIprefix}{doi:}
\providecommand{\ArXivprefix}{arXiv:}
\providecommand{\URLprefix}{URL: }
\providecommand{\Pubmedprefix}{pmid:}
\providecommand{\doi}[1]{\href{http://dx.doi.org/#1}{\path{#1}}}
\providecommand{\Pubmed}[1]{\href{pmid:#1}{\path{#1}}}
\providecommand{\bibinfo}[2]{#2}
\ifx\xfnm\relax \def\xfnm[#1]{\unskip,\space#1}\fi
\bibitem[{Dallas et~al.(2012)Dallas, Susarla, and Tan}]{Dallas2012}
\bibinfo{author}{T.~U.~T. Dallas}, \bibinfo{author}{A.~Susarla},
  \bibinfo{author}{Y.~Tan},
\newblock \bibinfo{title}{{Social Networks and the Diffusion of User- Generated
  Content : Evidence from YouTube Social Networks and the Diffusion of
  User-Generated Content : Evidence from YouTube}},
\newblock \bibinfo{journal}{Information Systems Research} \bibinfo{volume}{23}
  (\bibinfo{year}{2012}) \bibinfo{pages}{23--41}. \URLprefix
  \url{https://doi.org/10.1287/isre.1100.0339}.
  \href{http://arxiv.org/abs/isre.1100.0339}{{\tt arXiv:isre.1100.0339}}.
\bibitem[{Atefeh and Khreich(2015)}]{Atefeh2015}
\bibinfo{author}{F.~Atefeh}, \bibinfo{author}{W.~Khreich},
\newblock \bibinfo{title}{{A survey of techniques for event detection in
  Twitter}},
\newblock \bibinfo{journal}{Computational Intelligence} \bibinfo{volume}{31}
  (\bibinfo{year}{2015}) \bibinfo{pages}{133--164}. \URLprefix
  \url{https://doi.org/10.1111/coin.12017}.
  \href{http://arxiv.org/abs/arXiv:1011.1669v3}{{\tt arXiv:arXiv:1011.1669v3}}.
\bibitem[{Kaplan and Haenlein(2010)}]{Kaplan2010}
\bibinfo{author}{A.~M. Kaplan}, \bibinfo{author}{M.~Haenlein},
\newblock \bibinfo{title}{{Users of the world, unite! The challenges and
  opportunities of Social Media}},
\newblock \bibinfo{journal}{Business Horizons} \bibinfo{volume}{53}
  (\bibinfo{year}{2010}) \bibinfo{pages}{59--68}. \URLprefix
  \url{https://doi.org/10.1016/j.bushor.2009.09.003}.
  \href{http://arxiv.org/abs/1501.00994}{{\tt arXiv:1501.00994}}.
\bibitem[{Twitter(2018)}]{Twitter2018}
\bibinfo{author}{Twitter}, \bibinfo{title}{{Twitter is: what’s happening in
  the world and what people are talking about right now.}},
  \bibinfo{year}{2018}. \URLprefix \url{http://twitter.com/about}.
\bibitem[{Carlson(2011)}]{Carlson2011}
\bibinfo{author}{N.~Carlson}, \bibinfo{title}{{The Real History Of Twitter}},
  \bibinfo{year}{2011}. \URLprefix
  \url{http://www.businessinsider.com/how-twitter-was-founded-2011-4?op=1}.
\bibitem[{Kwak et~al.(2010)Kwak, Lee, Park, and Moon}]{Kwak2010}
\bibinfo{author}{H.~Kwak}, \bibinfo{author}{C.~Lee}, \bibinfo{author}{H.~Park},
  \bibinfo{author}{S.~Moon},
\newblock \bibinfo{title}{{What is Twitter, a Social Network or a News
  Media?}},
\newblock \bibinfo{journal}{Network}  (\bibinfo{year}{2010})
  \bibinfo{pages}{19--22}. \URLprefix
  \url{https://doi.org/10.1145/1772690.1772751}.
  \href{http://arxiv.org/abs/0809.1869v1}{{\tt arXiv:0809.1869v1}}.
\bibitem[{Stats(2018)}]{TwitterStats}
\bibinfo{author}{I.~L. Stats}, \bibinfo{title}{Twitter usage statistics},
  \bibinfo{year}{2018}. \URLprefix
  \url{http://www.internetlivestats.com/twitter-statistics/}.
\bibitem[{Bello-Orgaz et~al.(2016)Bello-Orgaz, Jung, and
  Camacho}]{Bello-Orgaz2016}
\bibinfo{author}{G.~Bello-Orgaz}, \bibinfo{author}{J.~J. Jung},
  \bibinfo{author}{D.~Camacho},
\newblock \bibinfo{title}{{Social big data: Recent achievements and new
  challenges}},
\newblock \bibinfo{journal}{Information Fusion} \bibinfo{volume}{28}
  (\bibinfo{year}{2016}) \bibinfo{pages}{45--59}. \URLprefix
  \url{https://doi.org/10.1016/j.inffus.2015.08.005}.
  \href{http://arxiv.org/abs/1505.06807}{{\tt arXiv:1505.06807}}.
\bibitem[{API(2018)}]{API2018}
\bibinfo{author}{T.~API}, \bibinfo{title}{{Getting Started | Twitter
  Developers}}, \bibinfo{year}{2018}. \URLprefix
  \url{https://dev.twitter.com/start}.
\bibitem[{Atefeh and Khreich(2015)}]{atefeh2015survey}
\bibinfo{author}{F.~Atefeh}, \bibinfo{author}{W.~Khreich},
\newblock \bibinfo{title}{A survey of techniques for event detection in
  twitter},
\newblock \bibinfo{journal}{Computational Intelligence} \bibinfo{volume}{31}
  (\bibinfo{year}{2015}) \bibinfo{pages}{132--164}.
\bibitem[{Aggarwal and Han(2014)}]{aggarwal2014frequent}
\bibinfo{author}{C.~C. Aggarwal}, \bibinfo{author}{J.~Han},
  \bibinfo{title}{Frequent pattern mining}, \bibinfo{publisher}{Springer},
  \bibinfo{year}{2014}.
\bibitem[{Ibrahim et~al.(2017)Ibrahim, Elbagoury, Kamel, and
  Karray}]{ibrahim2017tools}
\bibinfo{author}{R.~Ibrahim}, \bibinfo{author}{A.~Elbagoury},
  \bibinfo{author}{M.~S. Kamel}, \bibinfo{author}{F.~Karray},
\newblock \bibinfo{title}{Tools and approaches for topic detection from twitter
  streams: survey},
\newblock \bibinfo{journal}{Knowledge and Information Systems}
  (\bibinfo{year}{2017}) \bibinfo{pages}{1--29}.
\bibitem[{Zhang and Zhang(2002)}]{zhang2002association}
\bibinfo{author}{C.~Zhang}, \bibinfo{author}{S.~Zhang},
  \bibinfo{title}{Association rule mining: models and algorithms},
  \bibinfo{publisher}{Springer-Verlag}, \bibinfo{year}{2002}.
\bibitem[{Petkos et~al.(2014)Petkos, Papadopoulos, Aiello, Skraba, and
  Kompatsiaris}]{Petkos2014}
\bibinfo{author}{G.~Petkos}, \bibinfo{author}{S.~Papadopoulos},
  \bibinfo{author}{L.~Aiello}, \bibinfo{author}{R.~Skraba},
  \bibinfo{author}{Y.~Kompatsiaris},
\newblock \bibinfo{title}{{A soft frequent pattern mining approach for textual
  topic detection}},
\newblock in: \bibinfo{booktitle}{Proceedings of the 4th International
  Conference on Web Intelligence, Mining and Semantics (WIMS14) - WIMS '14},
  \bibinfo{year}{2014}, pp. \bibinfo{pages}{1--10}. \URLprefix
  \url{http://dl.acm.org/citation.cfm?doid=2611040.2611068}.
\bibitem[{{Huang, Jiajia and Peng} and Wang(2015)}]{HuangJiajiaandPeng2015}
\bibinfo{author}{M.~{Huang, Jiajia and Peng}}, \bibinfo{author}{Wang},
\newblock \bibinfo{title}{{Topic Detection from Large Scale of Microblog Stream
  with High Utility Pattern Clustering}},
\newblock \bibinfo{journal}{Proceedings of the 20th ACM SIGKDD international
  conference on Knowledge discovery and data mining}  (\bibinfo{year}{2015})
  \bibinfo{pages}{3--10}.
\bibitem[{Gaglio et~al.(2015)Gaglio, {Lo Re}, and Morana}]{Gaglio2015}
\bibinfo{author}{S.~Gaglio}, \bibinfo{author}{G.~{Lo Re}},
  \bibinfo{author}{M.~Morana},
\newblock \bibinfo{title}{{Real-time detection of twitter social events from
  the user's perspective}},
\newblock in: \bibinfo{booktitle}{IEEE International Conference on
  Communications}, volume \bibinfo{volume}{2015-Septe}, \bibinfo{year}{2015},
  pp. \bibinfo{pages}{1207--1212}.
\bibitem[{Choi and Park(2019)}]{Choi2019}
\bibinfo{author}{H.~J. Choi}, \bibinfo{author}{C.~H. Park},
\newblock \bibinfo{title}{{Emerging topic detection in twitter stream based on
  high utility pattern mining}},
\newblock \bibinfo{journal}{Expert Systems with Applications}
  \bibinfo{volume}{115} (\bibinfo{year}{2019}) \bibinfo{pages}{27--36}.
\bibitem[{Sayyadi et~al.(2009)Sayyadi, Hurst, and Maykov}]{Sayyadi2009}
\bibinfo{author}{H.~Sayyadi}, \bibinfo{author}{M.~Hurst},
  \bibinfo{author}{A.~Maykov},
\newblock \bibinfo{title}{{Event detection and tracking in social streams}},
\newblock in: \bibinfo{booktitle}{ICWSM - International Conference on Weblogs
  and Social Media}, \bibinfo{year}{2009}, pp. \bibinfo{pages}{311--314}.
  \href{http://arxiv.org/abs/1412.2188}{{\tt arXiv:1412.2188}}.
\bibitem[{Quercia et~al.(2012)Quercia, Askham, and Crowcroft}]{Quercia2012}
\bibinfo{author}{D.~Quercia}, \bibinfo{author}{H.~Askham},
  \bibinfo{author}{J.~Crowcroft},
\newblock \bibinfo{title}{{TweetLDA: Supervised Topic Classification and Link
  Prediction in Twitter}},
\newblock in: \bibinfo{booktitle}{WebSci}, \bibinfo{year}{2012}, pp.
  \bibinfo{pages}{1--4}.
\bibitem[{Liu and Qu(2012)}]{liu2012mining}
\bibinfo{author}{M.~Liu}, \bibinfo{author}{J.~Qu},
\newblock \bibinfo{title}{Mining high utility itemsets without candidate
  generation},
\newblock in: \bibinfo{booktitle}{Proceedings of the 21st ACM international
  conference on Information and knowledge management},
  \bibinfo{organization}{ACM}, \bibinfo{year}{2012}, pp.
  \bibinfo{pages}{55--64}.
\bibitem[{Saeed et~al.(2018)Saeed, Abbasi, Sadaf, Razzak, and
  Xu}]{saeed2018text}
\bibinfo{author}{Z.~Saeed}, \bibinfo{author}{R.~A. Abbasi},
  \bibinfo{author}{A.~Sadaf}, \bibinfo{author}{M.~I. Razzak},
  \bibinfo{author}{G.~Xu},
\newblock \bibinfo{title}{Text stream to temporal network-a dynamic heartbeat
  graph to detect emerging events on twitter},
\newblock in: \bibinfo{booktitle}{Pacific-Asia Conference on Knowledge
  Discovery and Data Mining}, \bibinfo{organization}{Springer},
  \bibinfo{year}{2018}, pp. \bibinfo{pages}{534--545}.
\bibitem[{Saeed et~al.(2019{\natexlab{a}})Saeed, Abbasi, Razzak, and
  Xu}]{saeed2019event}
\bibinfo{author}{Z.~Saeed}, \bibinfo{author}{R.~A. Abbasi},
  \bibinfo{author}{M.~I. Razzak}, \bibinfo{author}{G.~Xu},
\newblock \bibinfo{title}{Event detection in twitter stream using weighted
  dynamic heartbeat graph approach},
\newblock \bibinfo{journal}{arXiv preprint arXiv:1902.08522}
  (\bibinfo{year}{2019}{\natexlab{a}}).
\bibitem[{Saeed et~al.(2019{\natexlab{b}})Saeed, Abbasi, Razzak, Maqbool,
  Sadaf, and Xu}]{saeed2019enhanced}
\bibinfo{author}{Z.~Saeed}, \bibinfo{author}{R.~A. Abbasi},
  \bibinfo{author}{I.~Razzak}, \bibinfo{author}{O.~Maqbool},
  \bibinfo{author}{A.~Sadaf}, \bibinfo{author}{G.~Xu},
\newblock \bibinfo{title}{Enhanced heartbeat graph for emerging event detection
  on twitter using time series networks},
\newblock \bibinfo{journal}{Expert Systems with Applications}
  \bibinfo{volume}{136} (\bibinfo{year}{2019}{\natexlab{b}})
  \bibinfo{pages}{115--132}.
\bibitem[{Saeed et~al.(2020)Saeed, Abbasi, and Razzak}]{saeed2020evesense}
\bibinfo{author}{Z.~Saeed}, \bibinfo{author}{R.~A. Abbasi},
  \bibinfo{author}{I.~Razzak},
\newblock \bibinfo{title}{Evesense: What can you sense from twitter?},
\newblock in: \bibinfo{booktitle}{European Conference on Information
  Retrieval}, \bibinfo{organization}{Springer}, \bibinfo{year}{2020}, pp.
  \bibinfo{pages}{491--495}.
\bibitem[{Wo{\'z}niak et~al.(1995)Wo{\'z}niak, Gorzela{\'n}czyk, and
  Murakowski}]{wozniak1995two}
\bibinfo{author}{P.~Wo{\'z}niak}, \bibinfo{author}{E.~Gorzela{\'n}czyk},
  \bibinfo{author}{J.~Murakowski},
\newblock \bibinfo{title}{Two components of long-term memory.},
\newblock \bibinfo{journal}{Acta neurobiologiae experimentalis}
  \bibinfo{volume}{55} (\bibinfo{year}{1995}) \bibinfo{pages}{301--305}.
\bibitem[{Reimers and Gurevych(2019)}]{reimers2019sentence}
\bibinfo{author}{N.~Reimers}, \bibinfo{author}{I.~Gurevych},
\newblock \bibinfo{title}{Sentence-bert: Sentence embeddings using siamese
  bert-networks},
\newblock \bibinfo{journal}{arXiv preprint arXiv:1908.10084}
  (\bibinfo{year}{2019}).
\bibitem[{Asgari-Chenaghlu et~al.(2020)Asgari-Chenaghlu, Feizi-Derakhshi,
  Farzinvash, and Motamed}]{asgari2020multimodal}
\bibinfo{author}{M.~Asgari-Chenaghlu}, \bibinfo{author}{M.~R. Feizi-Derakhshi},
  \bibinfo{author}{L.~Farzinvash}, \bibinfo{author}{C.~Motamed},
\newblock \bibinfo{title}{A multimodal deep learning approach for named entity
  recognition from social media},
\newblock \bibinfo{journal}{arXiv preprint arXiv:2001.06888}
  (\bibinfo{year}{2020}).
\bibitem[{Zhao et~al.(2019)Zhao, Li, Zhang, Chiclana, and
  Viedma}]{zhao2019incremental}
\bibinfo{author}{Z.~Zhao}, \bibinfo{author}{C.~Li}, \bibinfo{author}{X.~Zhang},
  \bibinfo{author}{F.~Chiclana}, \bibinfo{author}{E.~H. Viedma},
\newblock \bibinfo{title}{An incremental method to detect communities in
  dynamic evolving social networks},
\newblock \bibinfo{journal}{Knowledge-Based Systems} \bibinfo{volume}{163}
  (\bibinfo{year}{2019}) \bibinfo{pages}{404--415}.
\bibitem[{Aiello et~al.(2013)Aiello, Petkos, Martin, Corney, Papadopoulos,
  Skraba, G{\"o}ker, Kompatsiaris, and Jaimes}]{aiello2013sensing}
\bibinfo{author}{L.~M. Aiello}, \bibinfo{author}{G.~Petkos},
  \bibinfo{author}{C.~Martin}, \bibinfo{author}{D.~Corney},
  \bibinfo{author}{S.~Papadopoulos}, \bibinfo{author}{R.~Skraba},
  \bibinfo{author}{A.~G{\"o}ker}, \bibinfo{author}{I.~Kompatsiaris},
  \bibinfo{author}{A.~Jaimes},
\newblock \bibinfo{title}{Sensing trending topics in twitter},
\newblock \bibinfo{journal}{IEEE Transactions on Multimedia}
  \bibinfo{volume}{15} (\bibinfo{year}{2013}) \bibinfo{pages}{1268--1282}.
\bibitem[{Sankaranarayanan et~al.(2009)Sankaranarayanan, Samet, Teitler,
  Lieberman, and Sperling}]{sankaranarayanan2009twitterstand}
\bibinfo{author}{J.~Sankaranarayanan}, \bibinfo{author}{H.~Samet},
  \bibinfo{author}{B.~E. Teitler}, \bibinfo{author}{M.~D. Lieberman},
  \bibinfo{author}{J.~Sperling},
\newblock \bibinfo{title}{Twitterstand: news in tweets},
\newblock in: \bibinfo{booktitle}{Proceedings of the 17th acm sigspatial
  international conference on advances in geographic information systems},
  \bibinfo{organization}{ACM}, \bibinfo{year}{2009}, pp.
  \bibinfo{pages}{42--51}.
\bibitem[{Phuvipadawat and Murata(2010)}]{phuvipadawat2010breaking}
\bibinfo{author}{S.~Phuvipadawat}, \bibinfo{author}{T.~Murata},
\newblock \bibinfo{title}{Breaking news detection and tracking in twitter},
\newblock in: \bibinfo{booktitle}{Web Intelligence and Intelligent Agent
  Technology (WI-IAT), 2010 IEEE/WIC/ACM International Conference on},
  volume~\bibinfo{volume}{3}, \bibinfo{organization}{IEEE},
  \bibinfo{year}{2010}, pp. \bibinfo{pages}{120--123}.
\bibitem[{Petrovi{\'c} et~al.(2010)Petrovi{\'c}, Osborne, and
  Lavrenko}]{petrovic2010streaming}
\bibinfo{author}{S.~Petrovi{\'c}}, \bibinfo{author}{M.~Osborne},
  \bibinfo{author}{V.~Lavrenko},
\newblock \bibinfo{title}{Streaming first story detection with application to
  twitter},
\newblock in: \bibinfo{booktitle}{Human Language Technologies: The 2010 Annual
  Conference of the North American Chapter of the Association for Computational
  Linguistics}, \bibinfo{organization}{Association for Computational
  Linguistics}, \bibinfo{year}{2010}, pp. \bibinfo{pages}{181--189}.
\bibitem[{Tembhurnikar and Patil(2015)}]{tembhurnikar2015topic}
\bibinfo{author}{S.~D. Tembhurnikar}, \bibinfo{author}{N.~N. Patil},
\newblock \bibinfo{title}{Topic detection using bngram method and sentiment
  analysis on twitter dataset}  (\bibinfo{year}{2015}) \bibinfo{pages}{1--6}.
\bibitem[{Osborne et~al.(2012)Osborne, Petrovic, McCreadie, Macdonald, and
  Ounis}]{osborne2012bieber}
\bibinfo{author}{M.~Osborne}, \bibinfo{author}{S.~Petrovic},
  \bibinfo{author}{R.~McCreadie}, \bibinfo{author}{C.~Macdonald},
  \bibinfo{author}{I.~Ounis},
\newblock \bibinfo{title}{Bieber no more: First story detection using twitter
  and wikipedia},
\newblock in: \bibinfo{booktitle}{SIGIR 2012 Workshop on Time-aware Information
  Access}, \bibinfo{year}{2012}.
\bibitem[{Cigarr{\'a}n et~al.(2016)Cigarr{\'a}n, Castellanos, and
  Garc{\'\i}a-Serrano}]{cigarran2016step}
\bibinfo{author}{J.~Cigarr{\'a}n}, \bibinfo{author}{{\'A}.~Castellanos},
  \bibinfo{author}{A.~Garc{\'\i}a-Serrano},
\newblock \bibinfo{title}{A step forward for topic detection in twitter: An
  fca-based approach},
\newblock \bibinfo{journal}{Expert Systems with Applications}
  \bibinfo{volume}{57} (\bibinfo{year}{2016}) \bibinfo{pages}{21--36}.
\bibitem[{Adedoyin-Olowe et~al.(2016)Adedoyin-Olowe, Gaber, Dancausa, Stahl,
  and Gomes}]{adedoyin2016rule}
\bibinfo{author}{M.~Adedoyin-Olowe}, \bibinfo{author}{M.~M. Gaber},
  \bibinfo{author}{C.~M. Dancausa}, \bibinfo{author}{F.~Stahl},
  \bibinfo{author}{J.~B. Gomes},
\newblock \bibinfo{title}{A rule dynamics approach to event detection in
  twitter with its application to sports and politics},
\newblock \bibinfo{journal}{Expert Systems with Applications}
  \bibinfo{volume}{55} (\bibinfo{year}{2016}) \bibinfo{pages}{351--360}.
\bibitem[{Dong et~al.(2017)Dong, Yang, Zhu, and Wang}]{dong2017discovering}
\bibinfo{author}{G.~Dong}, \bibinfo{author}{W.~Yang}, \bibinfo{author}{F.~Zhu},
  \bibinfo{author}{W.~Wang},
\newblock \bibinfo{title}{Discovering burst patterns of burst topic in
  twitter},
\newblock \bibinfo{journal}{Computers \& Electrical Engineering}
  \bibinfo{volume}{58} (\bibinfo{year}{2017}) \bibinfo{pages}{551--559}.
\bibitem[{Erra et~al.(2015)Erra, Senatore, Minnella, and
  Caggianese}]{erra2015approximate}
\bibinfo{author}{U.~Erra}, \bibinfo{author}{S.~Senatore},
  \bibinfo{author}{F.~Minnella}, \bibinfo{author}{G.~Caggianese},
\newblock \bibinfo{title}{Approximate tf--idf based on topic extraction from
  massive message stream using the gpu},
\newblock \bibinfo{journal}{Information Sciences} \bibinfo{volume}{292}
  (\bibinfo{year}{2015}) \bibinfo{pages}{143--161}.
\bibitem[{Becker et~al.(2011)Becker, Naaman, and Gravano}]{becker2011beyond}
\bibinfo{author}{H.~Becker}, \bibinfo{author}{M.~Naaman},
  \bibinfo{author}{L.~Gravano},
\newblock \bibinfo{title}{Beyond trending topics: Real-world event
  identification on twitter.},
\newblock \bibinfo{journal}{ICWSM} \bibinfo{volume}{11} (\bibinfo{year}{2011})
  \bibinfo{pages}{438--441}.
\bibitem[{Tang et~al.(2014)Tang, Wei, Yang, Zhou, Liu, and
  Qin}]{tang2014learning}
\bibinfo{author}{D.~Tang}, \bibinfo{author}{F.~Wei}, \bibinfo{author}{N.~Yang},
  \bibinfo{author}{M.~Zhou}, \bibinfo{author}{T.~Liu},
  \bibinfo{author}{B.~Qin},
\newblock \bibinfo{title}{Learning sentiment-specific word embedding for
  twitter sentiment classification.},
\newblock in: \bibinfo{booktitle}{ACL (1)}, \bibinfo{year}{2014}, pp.
  \bibinfo{pages}{1555--1565}.
\bibitem[{Hua et~al.(2016)Hua, Chen, Zhao, Lu, and
  Ramakrishnan}]{hua2016automatic}
\bibinfo{author}{T.~Hua}, \bibinfo{author}{F.~Chen}, \bibinfo{author}{L.~Zhao},
  \bibinfo{author}{C.-T. Lu}, \bibinfo{author}{N.~Ramakrishnan},
\newblock \bibinfo{title}{Automatic targeted-domain spatiotemporal event
  detection in twitter},
\newblock \bibinfo{journal}{GeoInformatica} \bibinfo{volume}{20}
  (\bibinfo{year}{2016}) \bibinfo{pages}{765--795}.
\bibitem[{Abdelhaq et~al.(2013)Abdelhaq, Sengstock, and
  Gertz}]{abdelhaq2013eventweet}
\bibinfo{author}{H.~Abdelhaq}, \bibinfo{author}{C.~Sengstock},
  \bibinfo{author}{M.~Gertz},
\newblock \bibinfo{title}{Eventweet: Online localized event detection from
  twitter},
\newblock \bibinfo{journal}{Proceedings of the VLDB Endowment}
  \bibinfo{volume}{6} (\bibinfo{year}{2013}) \bibinfo{pages}{1326--1329}.
\bibitem[{Lee and Sumiya(2010)}]{lee2010measuring}
\bibinfo{author}{R.~Lee}, \bibinfo{author}{K.~Sumiya},
\newblock \bibinfo{title}{Measuring geographical regularities of crowd
  behaviors for twitter-based geo-social event detection},
\newblock in: \bibinfo{booktitle}{Proceedings of the 2nd ACM SIGSPATIAL
  international workshop on location based social networks},
  \bibinfo{organization}{ACM}, \bibinfo{year}{2010}, pp.
  \bibinfo{pages}{1--10}.
\bibitem[{Li et~al.(2012)Li, Lei, Khadiwala, and Chang}]{li2012tedas}
\bibinfo{author}{R.~Li}, \bibinfo{author}{K.~H. Lei},
  \bibinfo{author}{R.~Khadiwala}, \bibinfo{author}{K.~C.-C. Chang},
\newblock \bibinfo{title}{Tedas: A twitter-based event detection and analysis
  system},
\newblock in: \bibinfo{booktitle}{Data engineering (icde), 2012 ieee 28th
  international conference on}, \bibinfo{organization}{IEEE},
  \bibinfo{year}{2012}, pp. \bibinfo{pages}{1273--1276}.
\bibitem[{Sakaki et~al.(2010)Sakaki, Okazaki, and
  Matsuo}]{sakaki2010earthquake}
\bibinfo{author}{T.~Sakaki}, \bibinfo{author}{M.~Okazaki},
  \bibinfo{author}{Y.~Matsuo},
\newblock \bibinfo{title}{Earthquake shakes twitter users: real-time event
  detection by social sensors},
\newblock in: \bibinfo{booktitle}{Proceedings of the 19th international
  conference on World wide web}, \bibinfo{organization}{ACM},
  \bibinfo{year}{2010}, pp. \bibinfo{pages}{851--860}.
\bibitem[{Snowsill et~al.(2010)Snowsill, Nicart, Stefani, De~Bie, and
  Cristianini}]{snowsill2010finding}
\bibinfo{author}{T.~Snowsill}, \bibinfo{author}{F.~Nicart},
  \bibinfo{author}{M.~Stefani}, \bibinfo{author}{T.~De~Bie},
  \bibinfo{author}{N.~Cristianini},
\newblock \bibinfo{title}{Finding surprising patterns in textual data streams},
\newblock in: \bibinfo{booktitle}{Cognitive Information Processing (CIP), 2010
  2nd International Workshop on}, \bibinfo{organization}{IEEE},
  \bibinfo{year}{2010}, pp. \bibinfo{pages}{405--410}.
\bibitem[{Cordeiro(2012)}]{cordeiro2012twitter}
\bibinfo{author}{M.~Cordeiro},
\newblock \bibinfo{title}{Twitter event detection: combining wavelet analysis
  and topic inference summarization},
\newblock in: \bibinfo{booktitle}{Doctoral symposium on informatics
  engineering}, \bibinfo{year}{2012}.

\end{thebibliography}

\end{document}